\documentclass{article}

\usepackage{arxiv,times}
\usepackage{natbib}
\renewcommand{\headeright}{arXiv preprint}
\renewcommand{\undertitle}{arXiv preprint}
\renewcommand{\shorttitle}{WorldOcc}
\usepackage[utf8]{inputenc}
\usepackage[T1]{fontenc}
\usepackage{hyperref}
\usepackage{url}
\usepackage{booktabs}
\usepackage{amsfonts}
\usepackage{amssymb}
\usepackage{amsmath}
\usepackage{nicefrac}
\usepackage{microtype}
\usepackage{subcaption}
\usepackage{float}
\usepackage{graphicx}
\usepackage{tikz}
\usetikzlibrary{calc}

\newlength{\cellh}
\newlength{\cellw}
\newlength{\cellgap}
\usepackage{xcolor}

  \newcommand{\rev}[1]{#1}
  \newcommand{\revisioncolor}{}

\usepackage{colortbl}
\usepackage{multirow}
\usepackage{algorithm}
\usepackage{algpseudocode}

\definecolor{rowlight}{RGB}{240,245,250}
\definecolor{ccgreen}{RGB}{0,150,60}
\definecolor{ccred}{RGB}{200,0,0}
\newcommand{\cmark}{\textcolor{ccgreen}{$\checkmark$}}
\newcommand{\xmark}{\textcolor{ccred}{$\times$}}
\newcommand{\rup}{\textcolor{red}{$\uparrow$}}
\newcommand{\rdown}{\textcolor{red}{$\downarrow$}}

\newcommand{\degree}{\ensuremath{{}^\circ}}

\newcommand{\occsq}[3]{\fcolorbox{black}[RGB]{#1,#2,#3}{{\color[RGB]{#1,#2,#3}\rule{0.45em}{0.45em}}}}
\newcommand{\occsqempty}{\fcolorbox{black}{white}{{\color{white}\rule{0.45em}{0.45em}}}}

\newcommand{\displabel}[3]{%
	\resizebox{\cellw}{!}{\scriptsize #1}\hspace{\cellgap}%
	\resizebox{\cellw}{!}{\scriptsize #2}\hspace{\cellgap}%
	\resizebox{\cellw}{!}{\scriptsize #3}}

\title{WorldOcc: A Real-Time Occupancy Simulator Towards
Closed-Loop Evaluation with Diverse and Learnable Traffic Behaviors}

\author{%
      Cong Feng\textsuperscript{1,2},
      Jiakai Wang\textsuperscript{2},
      Haozheng Wang\textsuperscript{3},
      Pu Feng\textsuperscript{2},\\
      \bfseries Xuesong Bai\textsuperscript{3},
      Hui He\textsuperscript{1},
      Xianglong Liu\textsuperscript{3}\\[6pt]
      \normalfont\small\textsuperscript{1}Harbin Institute of Technology\\
      \normalfont\small\textsuperscript{2}Zhongguancun Laboratory\\
      \normalfont\small\textsuperscript{3}Beihang University}

\makeatletter
\newcommand\blfootnote[1]{%
  \begingroup
  \renewcommand\thefootnote{}\footnotetext{#1}%
  \endgroup}
\makeatother

\hypersetup{
pdftitle={WorldOcc: A Real-Time Occupancy Simulator Towards Closed-Loop Evaluation with Diverse and Learnable Traffic Behaviors},
pdfsubject={autonomous driving, occupancy, closed-loop simulation},
pdfauthor={Cong Feng, Jiakai Wang, Haozheng Wang, Pu Feng, Xuesong Bai, Hui He, Xianglong Liu},
pdfkeywords={Occ3D, closed-loop, simulator, planning, learnable traffic behaviors},
}

\hypersetup{hidelinks}
\date{November 27, 2026}
\begin{document}
\maketitle
\blfootnote{Code will be released.}

\addtocontents{toc}{\protect\setcounter{tocdepth}{-1}}

\begin{abstract}
Closed-loop simulation for autonomous driving has garnered widespread attention
in recent years, yet, to our knowledge, no occupancy simulator is at once
real-time, high-fidelity, long-horizon, and able to exercise diverse, learnable
traffic.  This shortfall
stems largely from the prevailing practice of building occupancy simulators on a
learned world model, which entangles scene evolution with the decision making of
traffic participants (i.e., actors).  We
therefore rethink \emph{whether an occupancy simulator must rely on a learned
world model}, and propose WorldOcc, an occupancy simulator that is fast,
faithful, and sustained over long rollouts, without a learned world model.  At
its core is a \emph{policy-scene
disentangled architecture} that models the scene transition and the actor policy
independently, so each serves its own end, the transition for fidelity and the
policy for diversity and flexibility.  For high fidelity over a long horizon, we
design an \rev{\emph{error-bounded rigid-operator transition}}, Ego Warp-Recenter and Actor
Erase-Print, that reconstructs each occupancy state from its same-time recorded
reference rather than recursively forecasting voxel labels. For diverse
traffic behaviors, we tailor a \emph{plug-and-play actor policy network} that
learns and deploys behaviors, from normal to adversarial, effectively inside the simulator.
WorldOcc's ego transition runs at about $1270$\,FPS on
an RTX\,4090 GPU \rev{with pooled IoU $75.6\%$ over $16$\,s (the
log-replay ceiling is $100\%$)}.  Beyond simulation,
we introduce WorldOcc-Bench, to our knowledge the first closed-loop
benchmark for occupancy planners with diverse traffic, benchmarking six recent
planner families.  WorldOcc not only reveals the vulnerability of occupancy
planners under closed-loop conditions with diverse traffic, but also shows
potential as an effective post-training environment.
\end{abstract}

\begin{figure}[h]
	\centering
	\includegraphics[width=\textwidth]{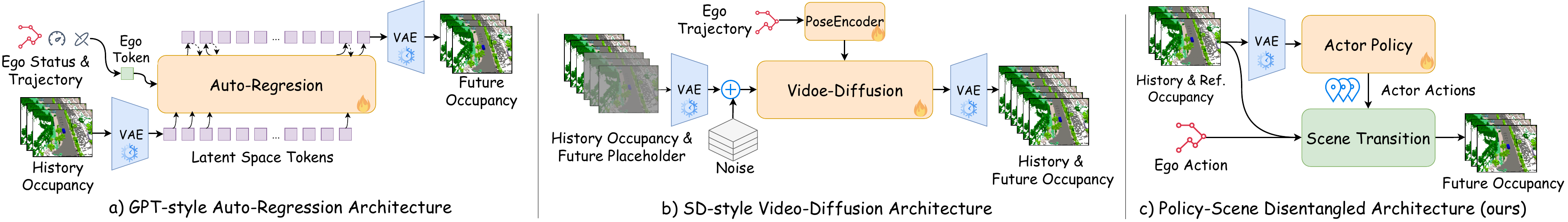}
	\caption{Occupancy simulators based on the learned world models, in
	(a) GPT-style \citep{gpt} autoregressive or (b) Stable Diffusion-style
	\citep{ldm} video-diffusion form.  Both entangle scene evolution and actor
	decision.  In contrast, (c) WorldOcc is a \emph{policy-scene disentangled
	architecture}, with an \rev{error-bounded rigid-operator transition} for the
	high-fidelity, long-horizon scene evolution and a plug-and-play actor
	policy network for the diverse, learnable traffic behaviors.}
	\label{fig:arch}
\end{figure}

\section{Introduction}
\label{sec:intro}

Closed-loop simulation for autonomous driving has drawn growing attention in
recent years.  Open-loop metrics such as L2 and collision are efficient but
correlate poorly with closed-loop metrics, while closed-loop evaluation remains
difficult to scale.  PDMS \citep{navsim} was therefore proposed as a
pseudo-closed-loop metric and is now widely adopted \citep{resworld}.  However, a recent study \citep{openloopcorr} finds that PDMS does not
always rank planners consistently with closed-loop evaluation, so
real-time closed-loop evaluation remains essential.  To obtain closed-loop
feedback, the community invests heavily in simulators.  Generative simulators,
such as DrivingSphere \citep{drivingsphere}
and DriveArena \citep{drivearena}, synthesize the future scene, and
reconstruction simulators, such as UniSim \citep{unisim} and
HUGSIM \citep{hugsim}, re-render a recorded scene from a driven pose.  Meanwhile, semantic
occupancy is a compelling representation for autonomous driving,
as it captures fine-grained 3D structure, can be obtained economically
from sparse LiDAR, and adapts to both vision and LiDAR modalities
\citep{occ3d,occworld}.  Yet, to the
best of our knowledge, no occupancy-modality closed-loop simulator is at once
real-time, high-fidelity, long-horizon, and able to exercise diverse, learnable
traffic.

Most prior work builds occupancy simulators with a learned world model that forecasts
the next occupancy under an ego condition, mainly as a GPT-style
autoregressive model such as I$^2$-World \citep{i2world} and OccWorld
\citep{occworld}, or a Stable
Diffusion-style video-diffusion model such as DOME \citep{dome} and OccSora
\citep{occsora} (Figure~\ref{fig:arch}).  Recent geometry-aware models improve this forecasting paradigm: GAST
\citep{gast} combines pose-driven BEV warping with learned spatial-temporal
refinement and reports both three-second and eight-second prediction settings.
Learned forecasting and reference-anchored simulation nevertheless serve
different purposes. Forecasting must infer unseen scene evolution from past
observations, whereas our simulator edits the recorded state at each time.
Moreover, occupancy forecasting does not by itself expose independently
trainable per-actor policies for controlled normal and adversarial traffic.
We therefore ask a more fundamental question: \emph{Must an occupancy
simulator rely on a learned world model?}

To answer this, we propose WorldOcc, a closed-loop occupancy simulator
that is fast, faithful, and sustained over long rollouts, without a learned
world model.  At its core is a \emph{policy-scene disentangled architecture}
that models the scene transition and the actor policy independently, so the
transition reuses recorded scene content while the policy controls diversity and
flexibility.  For high fidelity over a long
horizon, we design an \rev{\emph{error-bounded rigid-operator transition}} comprising Ego
Warp-Recenter and Actor Erase-Print, which reconstruct each occupancy state
from its same-time recorded reference. For diverse
traffic behaviors, we tailor a \emph{plug-and-play actor
policy network} that learns and deploys behaviors, from normal to
adversarial, effectively inside our simulator.  Harnessing our simulator, we take a step towards closed-loop
evaluation with diverse traffic, proposing WorldOcc-Bench on
Occ3D-nuScenes \citep{occ3d}, to our knowledge the first closed-loop benchmark
for occupancy planners.  Together, WorldOcc not only reveals
the vulnerability of occupancy planners under closed-loop conditions with diverse
traffic, but also shows potential as an effective post-training environment.

Our contributions are as follows.
\begin{itemize}
	\item We introduce WorldOcc, to the best of our knowledge the first
	occupancy-modality closed-loop simulator that is at once real-time,
	high-fidelity, and long-horizon, with diverse, learnable traffic behaviors,
	based on a \emph{policy-scene disentangled architecture} that separates scene
	evolution from actor behavior, without relying on a learned world model.
	\item \rev{We design an \emph{error-bounded rigid-operator transition} (Ego
	Warp-Recenter and Actor Erase-Print) that holds fidelity over a long horizon,
	and a \emph{plug-and-play actor policy network} that learns diverse
	traffic behaviors inside the simulator with much lower training costs.}
	\item We present WorldOcc-Bench, to our knowledge the first closed-loop
	benchmark for occupancy planners with diverse traffic, benchmarking six
	recent planner families.  Together, WorldOcc not only
	reveals the vulnerability of occupancy planners under closed-loop conditions
	with diverse traffic, but also shows potential as an effective post-training
	environment.
	\item Extensive experiments show that WorldOcc simulates in real time with
	high fidelity (pooled IoU $75.6\%$) over $16$\,s on
	Occ3D-nuScenes, with additional validation on Occ3D-Waymo,
	and that our actor policy network provides realistic and diverse traffic
	behaviors for our benchmark.
\end{itemize}

\section{Related Work}
\label{sec:related}

\subsection{Closed-Loop Simulators and Benchmarks for Autonomous Driving}
\label{sec:rel-sim}

\emph{Closed-loop Simulators.}  Reconstruction simulators re-render recorded
scenes into realistic observations but expose no learnable actor policy (UniSim
\citep{unisim}, HUGSIM \citep{hugsim}).  Game engine simulators such as CARLA
\citep{carla} provide reactive actors but synthetic observations, and generative
simulators such as DrivingSphere \citep{drivingsphere} and DriveArena
\citep{drivearena} synthesize realistic scenes with flexible controls but not in
real-time.  WorldOcc bridges the gap as a real-time closed-loop occupancy
simulator with diverse, learnable traffic behaviors.  \emph{Benchmarks of
Planners.}  CARLA-based suites (Bench2Drive \citep{bench2drive}, DriveE2E
\citep{drivee2e}) are closed-loop but keep the synthetic image modality.  NAVSIM
\citep{navsim} replays recordings but, as a short-horizon non-reactive
pseudo-simulation, cannot exercise the ego's responses to traffic.  NuPlan-R
\citep{nuplan-r} builds learned reactive traffic but only imitates recorded
behaviors.  In contrast, WorldOcc's plug-and-play actor policy can learn diverse
traffic behaviors inside the simulator.  Table~\ref{tab:pos} compares WorldOcc
with prior simulators and with closed-loop planning benchmarks along five
distinct axes.

\begin{table}[h]
	\centering
	\caption{\textbf{Comparison with closed-loop simulators and planning
	benchmarks.}  WorldOcc is a real-time occupancy simulator that supports
	diverse, learnable traffic behaviors.}
	\scriptsize
	\begin{tabular}{@{}llcccc@{}}
		\toprule
		\textbf{Method} & \textbf{Modality} & \textbf{Real-world} & \textbf{Long horizon} & \textbf{Real-time} & \textbf{Learnable Traffic Behavior}\\
		\midrule
		CARLA \citep{carla} & Image \& LiDAR & \xmark & \cmark & \cmark & \cmark\\
		UniSim \citep{unisim} & Image \& LiDAR & \cmark & \cmark & \xmark & \xmark\\
		NAVSIM \citep{navsim} & Image & \cmark & \xmark & \cmark & \xmark\\
		DriveArena \citep{drivearena} & Image & \cmark & \cmark & \xmark & \xmark\\
		DrivingSphere \citep{drivingsphere} & Image \& Occ. & \cmark & \cmark & \xmark & \xmark\\
		HUGSIM \citep{hugsim} & Image & \cmark & \cmark & \cmark & \xmark\\
		nuPlan-R \citep{nuplan-r} & Map \& State & \cmark & \cmark & \cmark & \cmark\\
		Bench2Drive \citep{bench2drive} & Image \& LiDAR & \xmark & \cmark & \cmark & \xmark\\
		DriveE2E \citep{drivee2e} & Image & \xmark & \cmark & \cmark & \xmark\\
		\rowcolor{rowlight}\textbf{WorldOcc} (ours) & Occ. & \cmark & \cmark & \cmark & \cmark\\
		\bottomrule
	\end{tabular}
	\label{tab:pos}
\end{table}

\subsection{Occupancy World Model for Planning, Simulation, and Scene Generation}
\label{sec:rel-occ}

Occupancy world models predict the evolution of the semantic scene, and mainly
fall into three lines based on their downstream applications.  The
planning line couples the forecast with driving, giving the ego a predictive
future to plan against (OccWorld \citep{occworld}, GEM \citep{gem}, OccTENS
\citep{occtens}); some additionally involve scene understanding and question
answering (OccLLaMA \citep{occllama}, Occ-LLM \citep{occllm}).  The simulation
line synthesizes future occupancy under an ego condition (DOME \citep{dome},
I$^2$-World \citep{i2world}, OccSora \citep{occsora}), with GAST
\citep{gast} additionally using geometric BEV warping and learned refinement
for occupancy forecasting; the scene generation
line generates occupancy data from layouts (AnyScene \citep{anyscene}, UniScene
\citep{uniscene}), \rev{from a one-frame static map with OccSim \citep{occsim}, a long-horizon
occupancy world-model-driven simulator,} or as large-scale 4D scenes
(DynamicCity \citep{dynamiccity}).  WorldOcc is an
occupancy simulator but \emph{not a learned world model}, providing a
closed-loop benchmark for planners while the underlying occupancy data can be
supplied by scene generators, bridging the two lines.

\section{Method}
\label{sec:method}

\subsection{Policy-Scene Disentangled Architecture}
\label{sec:overview}

\begin{figure*}[t]
	\centering
	\resizebox{\textwidth}{!}{\input{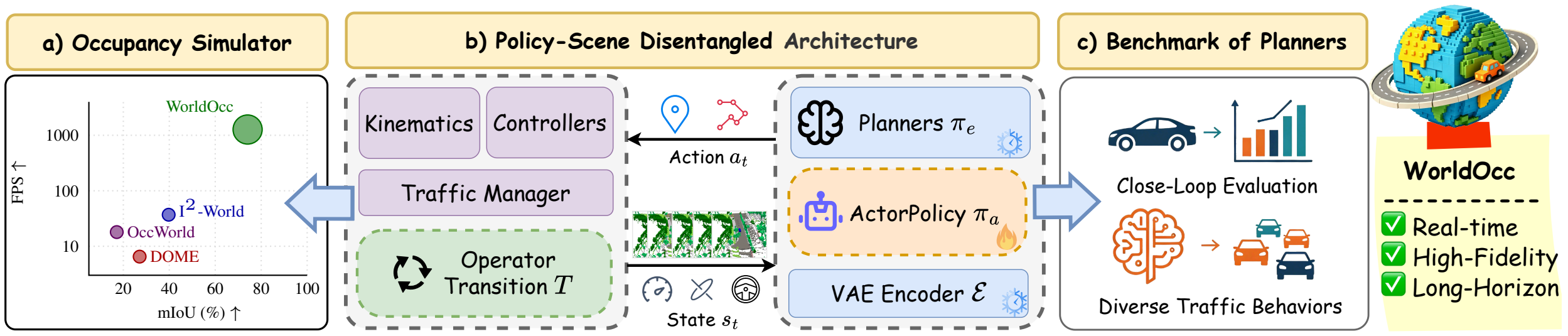}}
	\caption{\rev{\textbf{The WorldOcc Simulator.}}  (a) Occupancy fidelity (mIoU)
	vs.\ throughput (FPS) on Occ3D-nuScenes, measured on an RTX\,4090, with bubble
	area encoding the evaluated horizon. WorldOcc uses the pooled $16$\,s mIoU;
	the learned baselines use their $1$--$3$\,s average mIoU from Table~\ref{tab:fidelity}.
	WorldOcc FPS covers Ego Warp-Recenter only, excluding controller, actor
	editing, tokenizer and planner; baseline FPS follows the cited source.
	(b) Our policy-scene disentangled architecture (Section~\ref{sec:overview}).
	(c) We introduce a benchmark of planners with closed-loop evaluation and
	diverse traffic behaviors based on WorldOcc.}
	\label{fig:sim}
\end{figure*}

Prior world model simulators couple the static scene and the dynamic actors
into a single learned model, making both sub-optimal.  To
address this, we design a \emph{policy-scene
disentangled architecture} (Figure~\ref{fig:sim}b).  The occupancy scene is
advanced by the \rev{\emph{error-bounded rigid-operator transition}} of
Section~\ref{sec:transition}, while the controlled traffic
participants are governed by a plug-and-play actor policy network of
Section~\ref{sec:rl}.  First, we formulate the closed-loop simulation in
occupancy as a partially observable Markov decision process (POMDP)
\citep{pomdp},
$\langle S,A,T,R,\Omega,O,\gamma\rangle$, with state $s_t\in S$, action
$a_t\in A$, transition $T$, reward function $R$ with per-agent reward
$r^{(i)}_t\in\mathbb{R}$, observation $o_t\in\Omega$, observation function $O$,
and discount
$\gamma$.  At time $t$ the closed-loop state and an agent's observation are
\begin{equation}
	s_t=(X_t,E_t,G_t),\qquad o_t=O(s_t)=(B_t,\eta_t),
	\label{eq:state}
\end{equation}
where $X_t\in[0,C)^{H\times W\times Z}$ is the ego-centric semantic occupancy
scene of $H\times W\times Z$ voxels of cell size $\ell$, with $C$ classes, $B_t$
the agent's observation of $X_t$, and $\eta_t$ the agent's own state
($\eta_t=E_t$ for the ego,
$\eta_t=G_t$ for an actor).  $E_t$ and $G_t=\{G^{(i)}_t\}_{i=1}^{n}$ are the ego and collective actor
states, respectively (Appendix~\ref{app:simulator}).  The ego and the actors act
through their policies $\pi_e$ and $\pi_a$,
\begin{equation}
	a^{e}_t=\pi_e(o_t)=(w_{t+1},\ldots,w_{t+6})\in\mathbb{R}^{2\times6},\qquad
	a_t=\pi_a(o_t)=\{a^{(i)}_t\in\mathbb{R}^{2}\}_{i=1}^{n},
	\label{eq:actor}
\end{equation}
where $\pi_e$ is an \emph{external} planner under evaluation and $\pi_a$ a
\emph{plug-in} actor policy that the user can set (Section~\ref{sec:rl}).  The
ego action $a^{e}_t=(w_{t+1},\ldots,w_{t+6})\in\mathbb{R}^{2\times6}$ is a
six-waypoint trajectory with each waypoint $w_{t+1}\in\mathbb{R}^{2}$, following
the planning protocol of \citep{occworld}, and the
collective actor action $a_t=\{a^{(i)}_t\}_{i=1}^{n}$ collects each actor's aim
point $a^{(i)}_t\in\mathbb{R}^{2}$.  Unlike prior work
\citep{dome,i2world} that executes the actions (e.g., planned trajectory or
high-level command) directly in the simulator, we add kinematics
and controllers for more physically plausible traffic
behaviors (Appendix~\ref{app:simulator}).  The kinematics and controllers execute
the actions $a^{e}_t$ and $a_t$ into driven poses $E_{t+1}$ and $G_{t+1}$, so the transition
depends only on these poses and the current state $s_t$,
\begin{equation}
	s_{t+1}=T(s_t,E_{t+1},G_{t+1}).
	\label{eq:transition}
\end{equation}
We characterize our occupancy simulator by the pair $(T,\pi_a)$, where we model $T$
and $\pi_a$ independently: $T$ reuses recorded scene content, while
$\pi_a$ controls the behavior of selected traffic participants.

\subsection{\rev{Error-Bounded Rigid-Operator Transition}}
\label{sec:transition}

Prior world model-based simulators learn the transition function $T_\theta$
from data, but fail to maintain high fidelity over a long horizon.  Motivated by
NAVSIM \citep{navsim}, which evaluates planners by replaying real-world, recorded
scenes, we take the recorded states $s'_t$ and
$s'_{t+1}$ as \emph{reference states}.  We notice the large
\emph{spatial-temporal overlap of scene} (Table~\ref{tab:overlap}) and the
route-deviation termination used by the benchmark. These motivate using
$s'_{t+1}$ as a reference for constructing the closed-loop next state, so we
modify $T$ to operate on $s'_{t+1}$ instead,
\begin{equation}
	s_{t+1}=T\bigl(s'_{t+1},E_{t+1},G_{t+1}\bigr),
	\label{eq:transition-ref}
\end{equation}
where $s_{t+1}$ is the closed-loop next state.  The reference state
$s'_{t+1}=(X'_{t+1},E'_{t+1},G'_{t+1})$ supplies recorded scene content and
uncontrolled traffic with no re-driving overhead. It is a reference, not an
observed counterfactual outcome under the driven actions. The bound in
Appendix~\ref{app:overlap} is conditional on the stated pose deviations;
low open-loop error does not establish those conditions in closed loop. We
further decouple $T$ into Ego Warp-Recenter and Actor Erase-Print to execute
each kind of action.

\paragraph{Ego Warp-Recenter.}  We align the reference state $s'_{t+1}$ to
the driven ego pose.  Let $E'_t$ and $E'_{t+1}$ be the
reference ego poses at times $t$ and $t{+}1$, with positions $p'_t, p'_{t+1}$
and headings $\theta'_t, \theta'_{t+1}$, and let $E_{t+1}$ be the driven ego
pose at $t{+}1$, with position $p_{t+1}$ and heading
$\theta_{t+1}$.  We re-express the driven pose in the frame of the reference
state,
\begin{equation}
	\bar{p}_{t+1}=\mathrm{Rot}(\theta'_t-\theta'_{t+1})\,p_{t+1}
	+\mathrm{Rot}(-\theta'_{t+1})(p'_t-p'_{t+1}),
	\qquad
	\bar{\varphi}=\theta_{t+1}-(\theta'_{t+1}-\theta'_t),
	\label{eq:warp}
\end{equation}
where $\mathrm{Rot}(\alpha)$ is a rotation matrix for angle
$\alpha$, so $P=(\bar{p}_{t+1},\bar{\varphi})$ is the driven pose
re-expressed in the frame of the reference state.  For each driven-frame cell centre $c$, the implementation queries the
reference position $c'=\mathrm{Rot}(\bar{\varphi})c+\bar p_{t+1}$ and gathers
its semantic label from the containing reference voxel. With $b$ the grid's
lower metric corner, its index is $k'=\lfloor(c'-b)/\ell\rfloor$.
This inverse gathering avoids holes from forward splatting. Equivalently,
\begin{equation}
	c=\mathrm{Rot}(-\bar{\varphi})(c'-\bar{p}_{t+1}),\qquad
	X_{t+1}=\operatorname{Recenter}(X'_{t+1};\,P),
	\label{eq:recenter}
\end{equation}
Queries outside the reference grid receive the free label (Algorithm~\ref{alg:warp}). Unlike latent-space
warping, such as \citep{occsim,gast}, our operator can be viewed as an occupancy-space
warping.

\paragraph{Actor Erase-Print.}  We erase a real scene actor's
reference body and re-drive it under a novel behavior.  For a controlled
actor $i$, we set its own-class voxels inside its recorded bounding box at pose
$G'^{(i)}_{t+1}$ to the free class, giving the erased reference-state occupancy
$\bar{X}'_{t+1}$.  We
re-express the
actor's driven pose $G^{(i)}_{t+1}=(p^{(i)}_{t+1},\theta^{(i)}_{t+1})$ in the
driven ego coordinate, analogous to Eq.~\eqref{eq:warp},
\begin{equation}
	\bar{G}^{(i)}_{t+1}
	=\bigl(\mathrm{Rot}(-\bar{\varphi})(p^{(i)}_{t+1}-\bar{p}_{t+1}),\;
	\theta^{(i)}_{t+1}-\bar{\varphi}\bigr),
	\label{eq:print}
\end{equation}
The implementation prints the actor's captured body in the erased reference
grid, then applies Ego Warp-Recenter once to the edited grid,
\begin{equation}
	X_{t+1}
	=\operatorname{Recenter}\!\left(
        \bar{X}'_{t+1}\mathbin{\oplus}
        \Phi\big(P\circ\bar{G}^{(i)}_{t+1}\big);\,P\right),
	\label{eq:full}
\end{equation}
where $X_{t+1}$ is the next closed-loop occupancy in the driven ego frame,
$\operatorname{Recenter}(\cdot)$ is the operator defined by
Eq.~\eqref{eq:recenter}, and $\Phi(\cdot)$ places the actor's voxel body at the
given pose (Algorithm~\ref{alg:erase-print}). Here $P\circ\bar{G}^{(i)}$ maps
the actor pose back into the reference frame, and $\oplus$ denotes voxel
overwrite in the implementation's fixed actor order, not commutative set union.
Printing and nearest-neighbor gathering are not assumed to commute.

\subsection{Plug-and-Play Actor Policy Network}
\label{sec:rl}

Unlike prior approaches that predict a scene-policy coupled future, our actor
policy is plug-and-play, trained and deployed inside our simulator with no
finetuning of a heavy-weight world model.

\begin{figure*}[t]
	\centering
	\includegraphics[width=\textwidth]{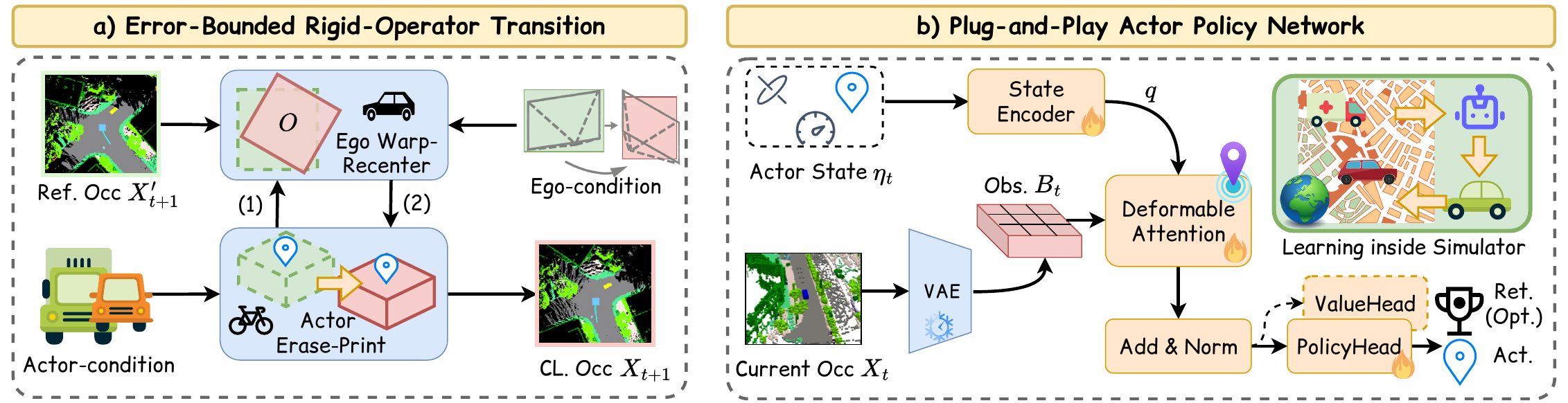}
	\caption{(a) The \rev{\emph{error-bounded rigid-operator transition}} comprising Ego Warp-Recenter and
	Actor Erase-Print using the next recorded state as a reference state, for
	reliable and trustworthy benchmarking.
	(b) The \emph{plug-and-play actor policy network} refines the actor query $q$ with
	a localized scene context from the observation $B_t$, and learns diverse
	traffic behaviors effectively.}
	\label{fig:tech}
\end{figure*}

\paragraph{Actor Policy Network.}
We first encode the occupancy state into a BEV observation of feature dimension
$d$ with a frozen
tokenizer (Figure~\ref{fig:tech}b) $\mathcal{E}$, and map each actor's
status $G^{(i)}_t$ to an actor query $q$ with an MLP,
\begin{equation}
	B_t=\mathcal{E}(X_t)\in\mathbb{R}^{H\times W\times d},
	\quad
	q=\mathrm{MLP}(G^{(i)}_t)\in\mathbb{R}^{d},
	\label{eq:tokens}
\end{equation}
We then refine $q$ with deformable attention
\citep{deformable_detr} over the scene $B_t$ around the actor's
current position $p^{(i)}_t$, so each decision depends on the actor's
localized perception of the environment,
\begin{equation}
	f_t = \operatorname{LayerNorm}\big(q + \operatorname{DeformAttention}(q, B_t;\, p^{(i)}_t)\big)
	     \in \mathbb{R}^{d},
	\label{eq:extractor}
\end{equation}
The feature $f_t$ feeds a policy head, and optionally a value head.  The action
space is the normalized next waypoint $a^{(i)}_t\in[-1,1]^2$, unnormalized by the
per-class displacement limit.  The network weights are shared across actors, so
at inference it can drive multiple controlled actors in a batch.  To augment our
simulator with natural and diverse interactions from traffic participants, we
instantiate the actor policy network on normal and adversarial behaviors,
detailed as follows.

\paragraph{Normal Traffic Behavior.}
Given that the underlying recorded data (i.e., nuScenes
\citep{nuscenes}) is predominantly benign, we obtain a normal behavior by
training the actor policy network to imitate recorded traffic
\citep{hussein2017imitation}.  We instantiate a deterministic policy head,
$a^{(i)}_t=\mu_\theta(f_t)$, supervised by the recorded next pose
$a^{\star(i)}_t$ in the current ego frame (Appendix~\ref{app:actor}).  We
minimize the $\ell_1$ behavior-cloning objective
\begin{equation}
	\mathcal{L}_{\mathrm{L1}}(\theta)=\mathbb{E}_t\big[\lVert
	\tanh(\mu_\theta(f_t))-a^{\star(i)}_t\rVert_1\big],
	\label{eq:bc-obj}
\end{equation}

\paragraph{Adversarial Traffic Behavior.}
Unlike a normal behavior, a
safety-critical adversarial behavior is rare in the benign recordings, so
imitating it would require collecting extra demonstration dataset.  We instead
learn the adversarial behavior directly from interactions inside our simulator
with deep reinforcement learning.  We pair a stochastic policy head with
a shared-feature value head,
\begin{equation}
	\pi(\cdot\mid o_t)=\mathcal{N}\big(\mu_\theta(f_t),\,\sigma_\theta\big),
	\qquad
	V_t=v_\phi(f_t),
	\label{eq:policy-ppo}
\end{equation}
where $\mu_\theta(f_t)$ and $\sigma_\theta$ are the parameterized Gaussian mean and standard
deviation, and $V_t=v_\phi(f_t)$ is the value head's estimate of the expected
discounted return $R^{(i)}_t=\sum_{k=0}^{\infty}\gamma^{k}r^{(i)}_{t+k}$.
We design an adversarial reward, consisting of a large reward on colliding with
the ego, a penalty for
leaving the sensing range, a distance-progress shaping toward the ego, and a
blocking penalty,
\begin{equation}
r^{(i)}_t =
\begin{cases}
	r_{\text{coll}}, & \text{the driven actor bbox intersects the ego (terminal),}\\
	r_{\text{out}}, & \text{the actor leaves the sensing range (terminal),}\\[2pt]
\lambda\,(d^{(i)}_t - d^{(i)}_{t+1}) - \kappa\,\beta^{(i)}_t, & \text{otherwise,}
\end{cases}
\label{eq:actor-reward}
\end{equation}
where $d^{(i)}_t=\lVert p^{(i)}_t\rVert$ is the actor-to-ego distance,
$r_{\text{coll}}>0$ and $r_{\text{out}}<0$ are the terminal collision reward and
out-of-range penalty, the
progress term $\lambda(d^{(i)}_t-d^{(i)}_{t+1})$ rewards closing the distance
with the ego,
$\beta^{(i)}_t\in[0,1]$ is the safety-braked fraction of the frame, and
$\lambda$, $\kappa$ weight progress and blocking.  We optimize with Proximal Policy
Optimization (PPO) \citep{ppo}.

\subsection{Closed-Loop Benchmark with Diverse Traffic}
\label{sec:eval}

Closed-loop evaluation with diverse traffic is essential for autonomous driving
\citep{bench2drive,drivee2e,li2026worldengine}.  We design \emph{WorldOcc-Bench}
on the Occ3D-nuScenes \citep{occ3d}, a closed-loop
benchmark in the occupancy modality with diverse traffic.  It evaluates a planner under three
traffic settings, namely (1) \emph{recorded},
where actors follow their recorded motion, (2) \emph{normal}, where the actors
are re-driven by the normal actor policy, and (3) \emph{adversarial}, where
actors are re-driven into safety-critical interactions with the adversarial
actor policy.

\paragraph{Traffic Manager.}
We tailor a \emph{range-aware spatial-temporal traffic manager} that selects
which actors to re-drive.  Motivated by nuPlan-R \citep{nuplan-r}, we sample
actors proximal to the ego with a per-class reachability range, and randomize
the frame where the policy starts to control them, so re-driven actors span
different types, positions, and timings.  We design two schedules:
\emph{spatial} installs one batch of $k$ actors at a single start frame for
concurrent pressure, while \emph{temporal} staggers $k$ actors over
independently drawn start frames for a sequence of separate encounters.
Sweeping $k$ along a per-schedule ladder exposes the actor load at which each
planner degrades.

\paragraph{Multi-Scenario Scoring.}
Because WorldOcc runs in real time, we can evaluate a planner on thousands of
scenarios within a reasonable time.  We therefore introduce a
\emph{multi-scenario} closed-loop score, \emph{WorldOcc-Score}, summarizing each
planner's performance under each setting and schedule.  Within a scenario, PDMS
\citep{navsim} averages a per-step component product over the $N$ driven frames
of that scenario,
\begin{equation}
	\text{PDMS} = \frac{1}{N}\sum_{t}
	\Big(\prod_{m\in\{\text{NC},\text{DAC}\}} m_t\Big)
	\cdot
	\frac{\sum_{m\in\{\text{TTC},\text{C}\}} w_m\,m_t}
	     {\sum_{m\in\{\text{TTC},\text{C}\}} w_m},
	\label{eq:pdms}
\end{equation}
where no collision (NC), drivable-area compliance (DAC), time-to-collision
(TTC), and comfort (C) score safety and ride quality, with $w_m$ the weight of
component $m$ in the weighted average over TTC and C, and route completion (RC)
is the fraction of the route driven.  Following \citep{hugsim}, we gate the PDMS to
prevent short driven routes from dominating the score, and aggregate over all
scenario cells $\mathcal{Q}_{S}$ within a traffic setting,
\begin{equation}
	\text{WorldOcc-Score}_S =
	\frac{1}{|\mathcal{Q}_{S}|}\sum_{j\in\mathcal{Q}_{S}}
	\text{PDMS}_{j}\cdot\text{RC}_{j},
	\label{eq:score}
\end{equation}
where $S \in \mathcal{T} = \{\text{recorded},\text{normal},\text{adversarial}\}$
is the traffic setting and $\mathcal{Q}_{S}$ contains its evaluated
(schedule, actor count, seed, scene) combinations. Each
(schedule, actor count, seed) cell receives equal weight and evaluates
150 scenes. The spatial ladder has three actor counts and the temporal
ladder five, so their weights within a re-driven traffic setting are
$3/8$ and $5/8$, respectively.  The recorded traffic setting
re-drives no actor, and has one deterministic cell, whereas the normal and
adversarial settings use both spatial and temporal schedules.  We report the score under
each traffic setting, and their mean as an overall score.

\section{Experiment Results}
\label{sec:experiments}

\subsection{Experimental Setup}
\label{sec:setup}

\textbf{Datasets.}  We evaluate our simulator on the Occ3D-nuScenes and
Occ3D-Waymo datasets \citep{occ3d}.
\rev{\textbf{Simulator Evaluation.}}  We compare occupancy fidelity in IoU/mIoU and
throughput in FPS following \citep{occworld}, over the $1$--$3$\,s horizon the
baselines cover, grouping them into two families, autoregressive
I$^2$-World \citep{i2world} and OccWorld \citep{occworld}, and video-diffusion
DOME \citep{dome} and DynamicCity \citep{dynamiccity}.
\rev{\textbf{Planner Benchmarking.}}  We benchmark occupancy-based and recent end-to-end
planners adapted to occupancy via the frozen I$^2$-tokenizer \citep{i2world}, namely GT-Following, OccWorld \citep{occworld}, Ego-MLP,
Go-Straight, the BEV-Planner family \citep{bevplanner}, DiffusionDrive
\citep{diffusiondrive}, ResWorld \citep{resworld}, SSR \citep{ssr}, and
WorldEngine (BC/RLFT) \citep{li2026worldengine}.  We also report reproduced
open-loop scores following \citep{bevplanner}.

\subsection{Implementation Details}
\label{sec:impl}

Our rigid-operator transition runs on both a
numba-JIT\footnote{\url{https://numba.pydata.org/}} CPU and a CUDA GPU.  The ego
plan is solved by an iLQR \citep{nuplan} once per occupancy frame and executed by
the bicycle model at a fixed $20$\,Hz; actors are driven by
an IDM \citep{idm}-inspired waypoint-following controller over a unicycle model.
Appendix~\ref{app:simulator} gives simulator details,
Appendix~\ref{app:actor} the actor-policy training details, and
Appendix~\ref{app:details} the benchmark details.

\subsection{A Real-Time, Long-Horizon, and High-Fidelity Simulator}

\begin{table}[t]
	\centering
	\caption{Occupancy fidelity on \textbf{Occ3D-nuScenes} in IoU/mIoU \rup (\%).
	$^{\dagger}$DynamicCity's conditional model. WorldOcc FPS measures the
	GPU-resident Ego Warp-Recenter operator; baseline FPS is reported by its source.}
	\label{tab:fidelity}
	\footnotesize
	\setlength{\tabcolsep}{3.5pt}%
	\resizebox{\textwidth}{!}{%
	\begin{tabular}{@{}l l ccc|cc c c@{}}
		\toprule
		\textbf{Method} & \textbf{Year} & 1\,s & 2\,s & 3\,s & 8\,s & 16\,s & \textbf{Avg} & \textbf{FPS}\\
		\midrule
		DOME \citep{dome} & 2024 & 43.99/35.11 & 35.36/25.89 & 29.74/20.29 & -- & -- & 36.36/27.10 & 6.5\\
		OccWorld \citep{occworld} & 2024 & 34.63/25.78 & 25.07/15.14 & 20.18/10.51 & -- & -- & 26.63/17.14 & 18.0\\
		I$^2$-World \citep{i2world} & 2025 & 54.29/47.62 & 49.43/38.58 & 45.69/32.98 & -- & -- & 49.80/39.73 & 37.0\\
		DynamicCity$^{\dagger}$ \citep{dynamiccity} & 2025 & 34.12/26.18 & 25.82/16.94 & -- & -- & -- & 30.0/21.6 & --\\
		\midrule
		\rev{\textbf{Log replay} (recorded frame, no warping)} & \rev{--} & \rev{100.0/100.0} & \rev{100.0/100.0} & \rev{100.0/100.0} & \rev{100.0/100.0} & \rev{100.0/100.0} & \rev{100.0/100.0} & \rev{--}\\
		\rowcolor{rowlight}\textbf{WorldOcc} (ours) & 2026 & 75.1/74.9 & 76.1/74.5 & 75.8/74.8 & 75.4/72.6 & 74.5/73.9 & \textbf{75.6/74.1} & \textbf{1270}\\
		\bottomrule
	\end{tabular}}
\end{table}

\rev{\textbf{Long-Horizon Fidelity.}}
We first measure long-horizon fidelity under recorded ego actions in
Table~\ref{tab:fidelity}. At the reported integer-second checkpoints
from $1$ to $16\,$s, WorldOcc holds IoU between $73\%$ and $80\%$, whereas the
learned world models decay over the $1$--$3$\,s they cover and stay far lower,
with I$^2$-World
dropping from $54\%$ to $46\%$ and DOME from $44\%$ to $30\%$.  \rev{The log-replay row returns the recorded frame itself, so its score is $100/100$; WorldOcc advances the simulated state and uses the recorded scene as the reference, so it is below $100\%$ by the pose-alignment cost.}  We present our
results on Occ3D-Waymo (Table~\ref{tab:waymo-sim}).  There we compare only with
I$^2$-World, as DOME and OccWorld report no Occ3D-Waymo results or checkpoints
and DynamicCity's conditional model is Occ3D-nuScenes only, so none of the three
matches our protocol.  WorldOcc holds
\rev{$91.7$--$94.2\%$ IoU at the reported integer-second horizons through $16\,$s at the native $10\,$Hz},
against I$^2$-World's zero-shot inference of $57.6$--$64.6\%$ over the
$1$--$3\,$s it covers.  The gap is similar under the $2\,$Hz protocol
\rev{($93.6$--$96.9\%$ vs.\ $46.9$--$58.3\%$)}.  \rev{Per-second values are in Table~\ref{tab:persecond}.}  \rev{At the $16\,$s horizon, 164/202 (10\,Hz) and 167/202 (2\,Hz) episodes contribute; episodes that terminate earlier do not contribute.}  We also compare the
teacher-forcing setting, the perception fidelity following
\citep{semcity}, and the counterfactual-maneuver fidelity in
Appendix~\ref{app:simulator}, showing the high-fidelity property of WorldOcc
comprehensively.  \looseness=-1 \rev{Voxel IoU is rotation-sensitive (Appendix~\ref{sub:app17}), so a small heading error can lower it without visible degradation.}

\begin{table}[t]
	\centering
	\caption{Occupancy fidelity on \textbf{Occ3D-Waymo} in IoU/mIoU \rup (\%).
	$^{\dagger}$I$^2$-World zero-shot.  DOME and OccWorld report no Occ3D-Waymo
	checkpoints, and DynamicCity is Occ3D-nuScenes only, so all three are
	omitted.}
	\label{tab:waymo-sim}
	\footnotesize
	\setlength{\tabcolsep}{3.5pt}%
	\resizebox{\textwidth}{!}{%
	\begin{tabular}{@{}l l ccc|cc c@{}}
		\toprule
		\textbf{Method} & \textbf{Freq} & 1\,s & 2\,s & 3\,s & 8\,s & 16\,s &
		\textbf{Avg}\\
		\midrule
		I$^2$-World$^{\dagger}$ \citep{i2world} & 2\,Hz & 58.29/42.84 & 51.84/35.58 &
		46.94/30.72 & -- & -- & 52.36/36.38\\
		\rowcolor{rowlight}\textbf{WorldOcc} (ours) & 2\,Hz & \rev{93.56/95.09} &
		\rev{93.61/94.98} & \rev{94.20/94.50} & \rev{95.11/96.02} & \rev{94.96/92.33} &
		\textbf{95.44/95.29}\\
		\midrule
		I$^2$-World$^{\dagger}$ \citep{i2world} & 10\,Hz & 64.58/49.35 & 60.71/43.03 &
		57.63/38.99 & -- & -- & 60.97/43.73\\
		\rowcolor{rowlight}\textbf{WorldOcc} (ours) & 10\,Hz & \rev{93.18/93.57} &
		\rev{92.29/90.99} & \rev{92.28/90.03} & \rev{93.53/92.68} & \rev{92.83/92.83} &
		\textbf{92.84/91.65}\\
		\bottomrule
	\end{tabular}}
\end{table}

\textbf{Throughput.}
On an RTX\,4090, WorldOcc advances the occupancy state at about $1270$\,FPS on
GPU for Ego Warp-Recenter alone (excluding controller, actor editing,
tokenizer, and planner inference), over thirty times I$^2$-World's reported $37.0$\,FPS, over seventy times
OccWorld's $18.0$\,FPS, and about two hundred times DOME's $6.5$\,FPS.  Even on
a CPU, WorldOcc still achieves $436$\,FPS.

\begin{figure}[h]
	\centering
	\includegraphics[width=0.75\textwidth]{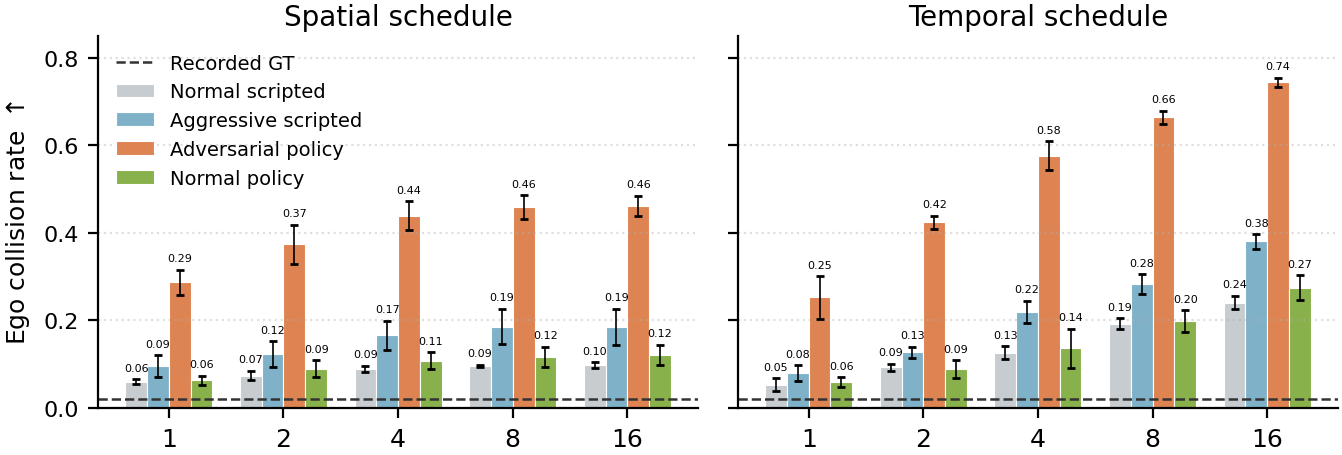}
	\caption{Ego collision rate vs.\ the number of controlled actors $k$, under
	the spatial and temporal schedules.  The dashed line
	is the recorded GT, which is invariant to the number of controlled actors.}
	\label{fig:actor-scaling}
\end{figure}

\textbf{Learned Actor Policies.}
\label{sec:actor-eval}
We evaluate the learned actor policies under the GT-Following ego against two
scripted behaviors, namely (1) \emph{normal scripted}, which re-drives each actor
along its recording, and (2) \emph{aggressive scripted}, which chases the ego,
over five seeds (Fig.~\ref{fig:actor-scaling}).  The \emph{adversarial} policy's
ego collision rate grows with the number of controlled actors $k$, reaching
$0.46$ spatially and $0.74$ temporally at $k{=}16$, above both scripted
baselines, while the \emph{normal} policy's rate of $0.125\pm0.067$ matches the
scripted re-drive's $0.111\pm0.058$, confirming that the actor policy network
can learn distinctive behaviors.  We further report the naturalness metrics and tracking accuracy of the normal
policy in Appendix~\ref{app:actor}, and summarize the training cost below,
showing efficiency and effectiveness of our policies.
\rev{\textbf{Lower Training Cost.}  The actor policy is light-weight: $91$K parameters
($78$K for BC), trained in $0.93$M PPO steps or $300$ BC episodes
(Table~\ref{tab:training-cost}); the PPO log records $925{,}184$ steps,
and evaluation uses the best checkpoint selected at $615{,}000$ steps.
Since the scene transition is parameter-free,
adding a behavior requires training only the actor policy.}

\begin{table}[H]
	\centering
	\revisioncolor
	\caption{\rev{Training cost of each method excluding tokenizers.
WorldOcc parameters count the active computation path (including the PPO critic);
the BC count excludes its critic and inactive status-only branch. External
budgets follow the cited papers/configuration and are not matched-compute
comparisons; unverified parameter counts and device counts are omitted.}}
	\label{tab:training-cost}
	\footnotesize
	\resizebox{\textwidth}{!}{%
	\begin{tabular}{@{}l l c c@{}}
		\toprule
		Method & Trained component & Params &
		Training compute\\
		\midrule
		DOME \citep{dome} & spatial-temporal diffusion & --- &
		$32\times$RTX\,4090, 2000 epochs\\
		OccWorld \citep{occworld} & spatial-temporal transformer & --- &
		$8\times$RTX\,4090, 200 epochs\\
		I$^2$-World \citep{i2world} & intra-inter scene encoder & --- &
		48 epochs\\
		\rowcolor{rowlight}WorldOcc (ours, RL) & actor policy network (+ critic) &
		91\,K & $1\times$RTX\,4090, 0.93\,M steps\\
		\rowcolor{rowlight}WorldOcc (ours, BC) & actor policy network &
		78\,K & $1\times$RTX\,4090, 300 episodes\\
		\bottomrule
	\end{tabular}}
\end{table}

\rev{GT-Following reaches 0.69 because the collision (NC, 0.73) and TTC
(0.70) terms dominate the product, while the lane, direction, and light terms
are 0.97--1.00.  The recorded replay is scored against recorded boxes at the
same timestamps, and we do not apply NAVSIM v2's human-violation filter, so this
is a NAVSIM-style variant.  For reference, NAVSIM v2 reports PDM-Closed at 51.3
EPDMS on navhard; the split and horizon differ, so the values are not directly
comparable.}

\subsection{The WorldOcc-Bench Leaderboard}
\label{sec:planner-results}

\textbf{Open-Loop vs. Closed-Loop Scores.}
Table~\ref{tab:planners} reports WorldOcc-Score across planners.  Every planner
falls well short of GT-Following, and the recorded-setting scores are nearly
uncorrelated with open-loop ADE (sample Pearson $r{=}+0.14$) and collision rate
($r{=}+0.12$ across nine variants; Fig.~\ref{fig:d1-correlation}), illustrating
differing rankings in this planner set \citep{bevplanner,navsim}. These sample
correlations are descriptive and sensitive to planner selection.  Our adapted ResWorld heads attain
open-loop L2 $0.35$/$0.32$\,m yet score poorly in closed loop
($0.034$/$0.049$), and Ego-MLP is near-best by open-loop ADE ($0.36$\,m) yet
scores $0.183$.  The recorded traffic setting alone is already a
challenge from self-conditioning covariate shift, since each planner is fed its
own \emph{driven} occupancy, leading to a poor RC of $0.07$--$0.33$.  Overall,
the benchmark presents a substantial challenge to current end-to-end and
occupancy-based planners.

\textbf{Vulnerability under Diverse Traffic.}
We further expose planners' vulnerability under diverse traffic behaviors.  The
\emph{normal} setting uses the learned behavior-cloned policy,
which imitates recorded motion, so most planners stay close to their recorded
score, with the perception-free Ego-MLP moving only $0.244\to0.241$.  The
\emph{adversarial} setting, driven by the reward-tuned policy, degrades nearly
every planner.  The occupancy-native OccWorld drops under adversarial ($0.193\to
0.191\to0.184$), so its failure is driven-occupancy drift rather than a
modality mismatch, while the perception-free Ego-MLP collapses only under
adversarial ($0.244\to0.241\to0.065$).  Even the GT-Following oracle falls from
$0.864$ to $0.804$ to $0.753$, so the traffic settings, not self-conditioning,
drive the degradation.  \rev{Appendix Fig.}~\ref{fig:rollout-a} shows the planner-by-traffic matrix
for one scene in the front-camera semantic view, where the GT-Following oracle
completes only the recorded clip and collides under both normal and adversarial
traffic, while Ego-MLP collides in all three settings, mirroring the scores of
Table~\ref{tab:planners}.

\textbf{Post-Training Environment.}
Beyond evaluation, the simulator also serves as a post-training environment.  We
port WorldEngine's trajectory-vocabulary planner \citep{li2026worldengine} and
follow their Reinforcement Learning Fine-Tuning (RLFT) recipe to fine-tune its
scoring head.  The planner scores a fixed $4096$-candidate vocabulary
against the occupancy latent, and one-step RLFT lifts the recorded score from
$0.273$ to $0.293$, the best after the GT-Following oracle.  We further study
multi-step RLFT beyond WorldEngine's setting and find the WorldOcc-Score peaks
at $0.309$ with two steps (Appendix~\ref{app:details}); long-horizon RLFT on a
driving policy is left to future work.

\begin{table}[h]
	\centering
	\caption{WorldOcc-Bench results with reproduced open-loop results.
	$^{\dagger}$GT-Following is a human reference rather than a learned planner.}
	\resizebox{\textwidth}{!}{%
	\begin{tabular}{@{}l ccccc cccc cc@{}}
		\toprule
		& \multicolumn{5}{c}{\textbf{PDMS subscores}} & \multicolumn{4}{c}{\textbf{WorldOcc-Score}} & \multicolumn{2}{c}{\textbf{Open-loop}}\\
		\cmidrule(r){2-6}\cmidrule(r){7-10}\cmidrule(l){11-12}
		\textbf{Baseline} & \textbf{NC} \rup & \textbf{DAC} \rup & \textbf{TTC} \rup & \textbf{Comf.} \rup & \textbf{RC} \rup & \textbf{Recorded} \rup & \textbf{Normal} \rup & \textbf{Adversarial} \rup & \textbf{Overall} \rup & \textbf{L2 (m)} \rdown & \textbf{Coll. (\%)} \rdown\\
		\midrule
\textbf{GT-Following}$^{\dagger}$ & 0.992 & 0.913 & 0.991 & 0.871 & 0.937 & 0.864 & 0.804 & 0.753 & 0.807 & 0.00 & 0.00\\
\rowcolor{rowlight}\textbf{OccWorld} \citep{occworld} & 0.953 & 0.875 & 0.946 & 0.740 & 0.254 & 0.193 & 0.191 & 0.184 & 0.189 & 0.63 & 3.18\\
\rowcolor{rowlight}\textbf{Ego-MLP} \citep{bevplanner} & 0.966 & 0.781 & 0.957 & 0.910 & 0.286 & 0.244 & 0.241 & 0.065 & 0.183 & 0.36 & 0.56\\
\rowcolor{rowlight}\textbf{Go-Straight} \citep{bevplanner} & 0.963 & 0.793 & 0.962 & 1.000 & 0.254 & 0.227 & 0.224 & 0.221 & 0.224 & 0.55 & 1.05\\
\textbf{BEV-Planner+} \citep{bevplanner} & 0.963 & 0.788 & 0.957 & 0.865 & 0.304 & 0.251 & 0.246 & 0.241 & 0.246 & 0.34 & 0.50\\
\rowcolor{rowlight}\textbf{BEV-Planner++} \citep{bevplanner} & 0.966 & 0.797 & 0.961 & 0.930 & 0.296 & 0.252 & 0.249 & 0.242 & 0.248 & 0.37 & 0.79\\
\textbf{DiffusionDrive} \citep{diffusiondrive} & 0.944 & 0.792 & 0.937 & 0.797 & 0.288 & 0.249 & 0.250 & 0.228 & 0.242 & 0.32 & 0.42\\
\textbf{ResWorld} \citep{resworld} & 0.922 & 0.696 & 0.905 & 0.610 & 0.070 & 0.034 & 0.035 & 0.035 & 0.034 & 0.35 & 0.40\\
\rowcolor{rowlight}\textbf{ResWorld$\lozenge$} \citep{resworld} & 0.905 & 0.715 & 0.886 & 0.506 & 0.093 & 0.057 & 0.045 & 0.045 & 0.049 & 0.32 & 0.38\\
\textbf{SSR} \citep{ssr} & 0.959 & 0.761 & 0.952 & 0.790 & 0.255 & 0.214 & 0.212 & 0.196 & 0.207 & 0.34 & 0.57\\
\rowcolor{rowlight}\textbf{WorldEngine (BC)} \citep{li2026worldengine} & 0.941 & 0.874 & 0.930 & 0.791 & 0.312 & 0.273 & 0.268 & 0.247 & 0.263 & --- & ---\\%
\textbf{WorldEngine (RLFT)} \citep{li2026worldengine} & \textbf{0.976} & \textbf{0.970} & \textbf{0.969} & 0.960 & \textbf{0.331} & \textbf{0.293} & \textbf{0.287} & \textbf{0.263} & \textbf{0.281} & --- & ---\\%
		\bottomrule
	\end{tabular}}
	\label{tab:planners}
\end{table}

\section{\rev{Further Analyses}}
\label{sec:analyses}

\begingroup\revisioncolor
\paragraph{Traffic-Rule Layer.}
\looseness=-1 We extend WorldOcc to nuPlan-Occ~\citep{uniscenev2}, a large-scale
occupancy dataset built on nuPlan~\citep{nuplan} whose map supplies lane
boundaries and whose logs carry per-frame traffic-light states, neither of which
Occ3D-nuScenes provides.  On its val split, replaying recorded scenes keeps the
occupancy state consistent with the reference frames, with lane classes aligned to
the nuPlan boundaries (lane IoU $0.83$--$0.87$; Fig.~\ref{fig:nuplan-occ};
Table~\ref{tab:nuplan-occ}).  A rule layer then reads those signals and bounds the
planner's speed reference, and we score the rollouts with the terms of
Table~\ref{tab:nuplan-rules}, which follow NAVSIM v2~\citep{navsimv2}:
traffic-light compliance (TLC), driving-direction compliance (DDC), no at-fault
collision (NC), lane keeping (LK), and the Extended Predictive Driver Model Score
(EPDMS).  Applied to
two planners, the terms separate a compliant recorded replay from a constant-speed
policy that ignores the lights.

\begin{table}[h]
	\centering
	\revisioncolor
	\caption{\rev{Traffic Rule Layer score terms on the val split of NuPlan-Occ mini~\citep{uniscenev2}.}}
	\label{tab:nuplan-rules}
	\footnotesize
	\begin{tabular}{@{}l ccccc@{}}
		\toprule
		\textbf{Policy} & \textbf{TLC} & \textbf{DDC} & \textbf{NC} & \textbf{LK} & \textbf{EPDMS}\\
		\midrule
		GT-Following (recorded replay) & 0.97 & 0.97 & 0.73 & 1.00 & 0.69\\
		Go-Straight & 0.67 & 0.80 & 0.10 & 0.07 & 0.09\\
		\bottomrule
	\end{tabular}
\end{table}

\paragraph{Scene-Generator Composition.}
\rev{We also explore whether a pre-trained generator can replace the recorded
stream.  Ten DynamicCity samples~\citep{dynamiccity} retain high reference-relative
fidelity after Warp-Recenter; we visualize each reference sample at four timestamps
(Fig.~\ref{fig:compose-main}), and we score the composed rollouts against the
reference stream (Table~\ref{tab:gen-compose}), where the fidelity gap to the
recorded stream is under $0.02$\,mIoU.}

\paragraph{Adversarial-Policy Realism.}
We further evaluate the adversarial policy by counting the collisions it causes, classifying them via the
four coarse kinematic families inspired by the
NHTSA pre-crash typology~\citep{najm2007precrash}, and measuring the kinematics
of the evasive motion.  Over 70 interaction events, the policy produces 42 collisions that
cover four families (Fig.~\ref{fig:conflict-bars}).  The controlled actors have
off-road and non-ego conflict densities close
to the scripted-aggressive baseline (Fig.~\ref{fig:actor-3metrics}), and we report the
driven-motion kinematics (Appendix
Fig.~\ref{fig:conflict-examples}; Tables~\ref{tab:interaction-stats}
and~\ref{tab:conflict-kinematics}).

\begin{figure}[h]
	\centering
	\includegraphics[width=\textwidth]{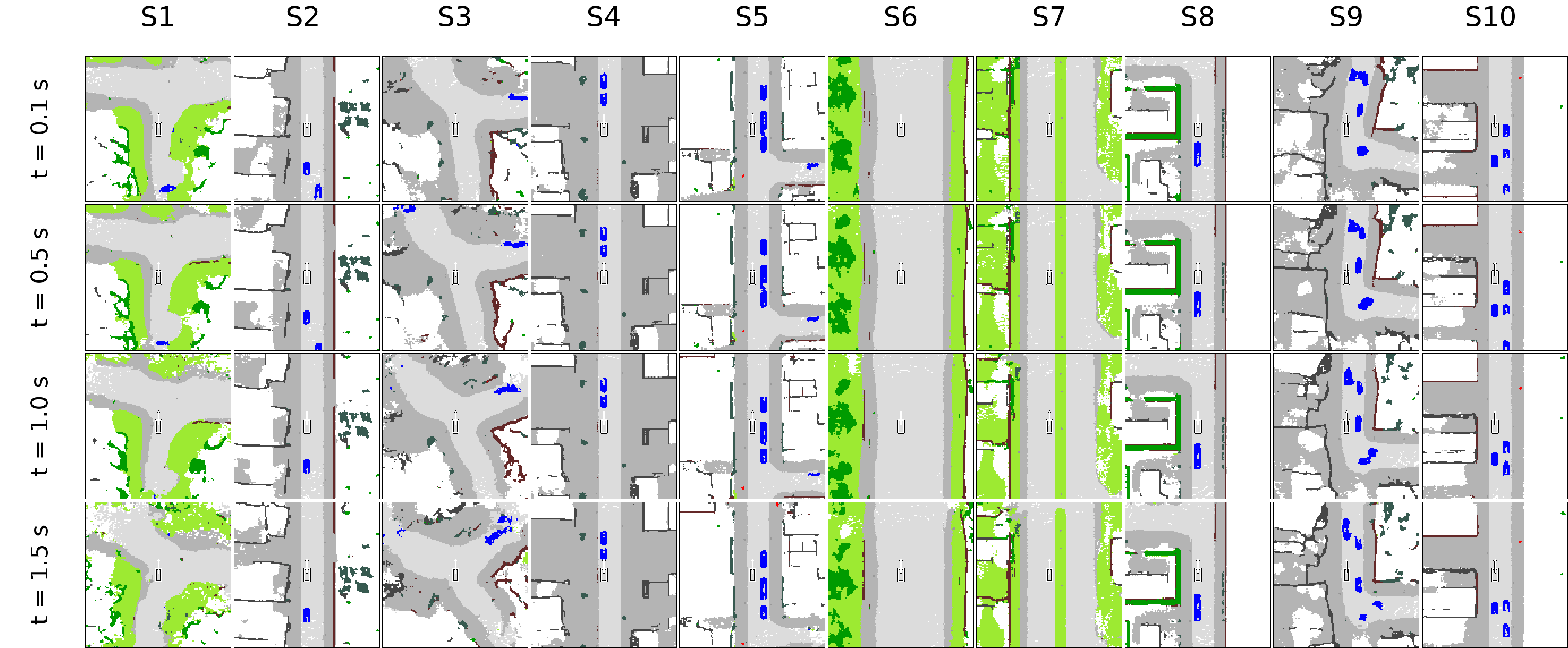}
	\revisioncolor
	\caption{\rev{Composition with DynamicCity~\citep{dynamiccity} on the
	CARLA-SC dataset~\citep{motionsc}.  Columns are the ten samples and rows
	are timestamps at $0.1$, $0.5$, $1.0$ and $1.5$\,s.}}
	\label{fig:compose-main}
\end{figure}

\begin{figure}[h]
	\centering
	\begin{subfigure}[b]{0.50\textwidth}
		\centering
		\includegraphics[width=\linewidth]{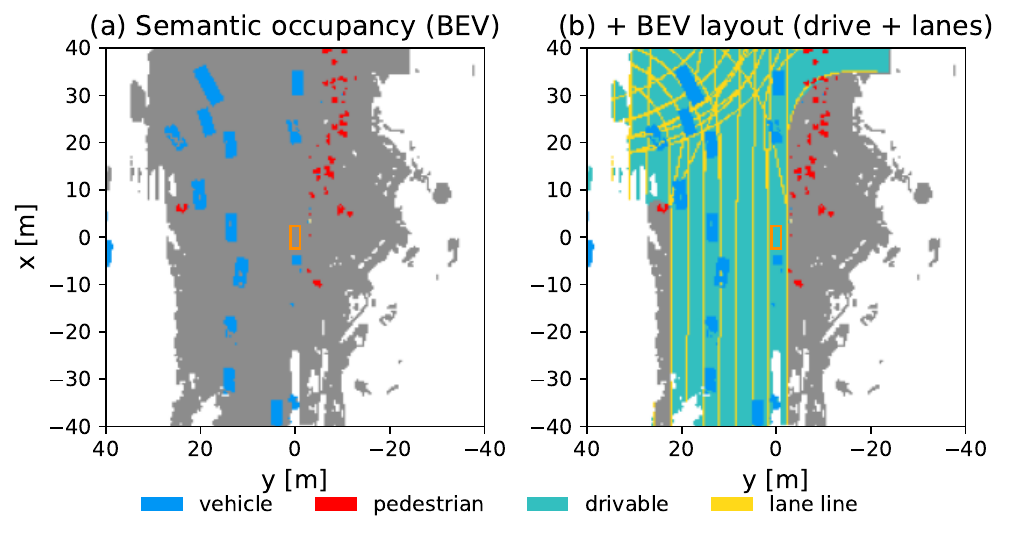}
		\caption{\rev{NuPlan-Occ}}
		\label{fig:nuplan-occ}
	\end{subfigure}\hfill
	\begin{subfigure}[b]{0.46\textwidth}
		\centering
		\includegraphics[width=\linewidth]{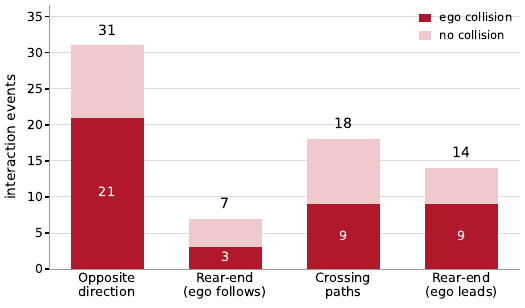}
		\caption{\rev{Conflict families}}
		\label{fig:conflict-bars}
	\end{subfigure}
	\revisioncolor
	\caption{\rev{NuPlan-Occ layout and conflict-family counts.  (a) Semantic
	occupancy in the BEV with projected nuPlan lanes and drivable surface.
	(b) The 70 interaction events of the learned adversary, grouped into four
	kinematic families; dark segments are ego collisions (42 total).}}
	\label{fig:analysis}
\end{figure}

\endgroup

\section{Conclusion}
\label{sec:conclusion}

In this paper, we introduced WorldOcc, a real-time, high-fidelity,
long-horizon occupancy simulator that avoids a learned world model.  Our
policy-scene disentangled architecture separates scene evolution from
actor behavior, so the transition serves fidelity and the policy serves
diversity.  Harnessing it, we propose WorldOcc-Bench, a closed-loop benchmark
for occupancy planners with diverse traffic,
exposing planner vulnerability and showing potential as an effective
post-training environment.

\textbf{Limitations and Future Work.}
\rev{We evaluate in occupancy space only and do not run a sensor-level
evaluation, where the occupancy would first be rendered to camera or LiDAR
through pipelines such as UniScene or AnyScene.  The next step we plan to take is
to close that gap, rendering the simulated occupancy and evaluating perception and
planning on the rendered sensors.}

\subsection*{AI use statement}

We used generative AI tools to assist with drafting, editing, code maintenance,
and reproducibility checks. Quantitative results are computed by executable
code, and we take responsibility for the methods, data, results, and all content.

\subsection*{Reproducibility statement}

All experiments use public datasets.
The operators, kinematics, controllers, benchmark protocol, and metrics are
specified in the main text and appendices.  OccWorld uses its released checkpoint; occupancy-adapted planners use
checkpoints trained in this project. The reproduction guide records configurations,
checkpoint hashes, validation coverage, and remaining verification gaps. Code is
prepared for release; experimental archives and weights are supplied separately.
Rerun records distinguish the original measurements from results on other
hardware and software stacks; floating-point inference can change long-horizon
planner scores, so bitwise equality across environments is not assumed.

\subsection*{Ethics statement}

This work uses the public Occ3D dataset and the published planners, simulators.
No human subjects or personal
data are involved, and the simulator is used only for offline evaluation and RL
training.

\bibliographystyle{plainnat}
\bibliography{references}

\appendix

\section*{Appendix}
In this appendix, we supplement the following materials to support the findings
and conclusions drawn in the main body of this paper.

\addtocontents{toc}{\protect\setcounter{tocdepth}{2}}
\tableofcontents

\section{Operator Transition Details}
\label{app:overlap}
\label{sub:app1}

We study empirical fidelity and a conditional, reference-relative label-difference
bound for the proposed rigid-operator transition. The transition $T$ of Section~\ref{sec:transition}
reduces to two operators, Ego Warp-Recenter and Actor Erase-Print, justified by
two observations.  Adjacent frames share most of their content, and the
benchmark terminates when driven-route deviation exceeds its threshold.
The $0.32$--$0.63$\,m open-loop ADE (Table~\ref{tab:openloop}) is descriptive
evidence on recorded inputs, not a bound on executed trajectories. The analysis
below concerns differences from a same-time recorded grid.

\subsection{Spatial-Temporal Overlap of Scene}
\label{sub:app2}
We verify the first observation empirically.  On the whole validation set, we warp each
frame-$t$ occupancy into the frame-$(t{+}1)$ ego frame with the recorded ego
motion and score the warped frame against the GT next-frame occupancy.
Table~\ref{tab:overlap} reports the per-class IoU with
no warp, with Ego Warp-Recenter alone, and with Actor Erase-Print added.

\begin{table}[h]
	\centering
	\caption{Spatial-temporal overlap of consecutive frames.  mIoU averages the
	$17$ semantic classes, excluding free.  IoU is the
	binary non-free occupancy IoU.}
	\label{tab:overlap}
	\resizebox{\textwidth}{!}{%
	\begin{tabular}{@{}lcccccccccccccccccccc@{}}
		\toprule
		& \rotatebox{90}{\occsq{50}{120}{255}\;others} & \rotatebox{90}{\occsq{203}{192}{255}\;barrier} & \rotatebox{90}{\occsq{0}{255}{255}\;bicycle} & \rotatebox{90}{\occsq{245}{150}{0}\;bus} & \rotatebox{90}{\occsq{255}{255}{0}\;car} & \rotatebox{90}{\occsq{0}{127}{255}\;constr.} & \rotatebox{90}{\occsq{0}{0}{255}\;motor.} & \rotatebox{90}{\occsq{150}{240}{255}\;pedestr.} & \rotatebox{90}{\occsq{0}{60}{135}\;cone} & \rotatebox{90}{\occsq{240}{32}{160}\;trailer} & \rotatebox{90}{\occsq{255}{0}{255}\;truck} & \rotatebox{90}{\occsq{137}{137}{139}\;drivable} & \rotatebox{90}{\occsq{75}{0}{75}\;o.flat} & \rotatebox{90}{\occsq{80}{240}{150}\;sidewalk} & \rotatebox{90}{\occsq{250}{230}{230}\;terrain} & \rotatebox{90}{\occsq{0}{175}{0}\;manmade} & \rotatebox{90}{\occsq{127}{255}{0}\;veget.} & \rotatebox{90}{\occsqempty\;free} & \rotatebox{90}{\textbf{mIoU} \rup} & \rotatebox{90}{\textbf{IoU} \rup}\\
		\midrule
		\textbf{No warp} (\%) & 13.3 & 25.2 & 8.9 & 31.8 & 21.5 & 17.5 & 8.4 & 7.7 & 9.3 & 19.0 & 21.0 & 49.4 & 33.7 & 32.7 & 30.5 & 19.0 & 18.8 & 94.5 & \textbf{21.6} & \textbf{30.5}\\
		\textbf{Warp} (\%) & 59.6 & 62.5 & 40.5 & 46.0 & 39.3 & 60.6 & 32.0 & 30.0 & 52.1 & 52.8 & 49.0 & 59.1 & 51.3 & 52.7 & 49.6 & 59.3 & 60.6 & 97.4 & \textbf{50.4} & \textbf{59.1}\\
		\rowcolor{rowlight}\textbf{Warp + Print} (\%) & 59.6 & 62.5 & 43.9 & 65.2 & 57.1 & 58.0 & 48.9 & 38.6 & 52.1 & 55.4 & 63.2 & 59.2 & 51.3 & 52.7 & 49.6 & 59.3 & 60.6 & 97.5 & \textbf{55.1} & \textbf{60.0}\\
		\bottomrule
	\end{tabular}}
\end{table}

Warping nearly doubles the non-empty IoU, from 30.5\% to 59.1\%, with free
space at 97.4\%.  The residual concentrates on dynamic objects such as car
39.3\%, motorcycle 32.0\% and pedestrian 30.0\%, and grows with ego
displacement.  82\% non-empty IoU below 1\,m of motion vs 53\% above 4\,m.  Actor Erase-Print on the
warp repositions each actor at its next-frame pose and recovers part of this
residual, lifting mIoU from 50.4\% to 55.1\%.  Car 39.3\%$\to$57.1\%, bus
46.0\%$\to$65.2\%, truck 49.0\%$\to$63.2\%, motorcycle 32.0\%$\to$48.9\%, and
pedestrian 30.0\%$\to$38.6\%.  Static structure is largely shared.  Buildings
59\%, roads 59\%, vegetation 61\%.  Thus Ego
Warp-Recenter alone nearly doubles overlap (30.5\% to 59.1\% IoU) and Actor
Erase-Print recovers the residual, to 60.0\%.  In
closed-loop GT replay the occupancy state is re-centred from the recorded next frame,
reaching 75.6\% IoU.  Appendix~\ref{app:simulator} reports the per-operator
timings.

\begin{algorithm}[h]
	\caption{Ego Warp-Recenter transition, with $X'_{t+1}$ the reference occupancy, $P$ the driven pose, and $\ell$ the cell size}
	\label{alg:warp}
	\begin{algorithmic}[1]
		\State $X_{t+1}\leftarrow\textsc{Free}(H\times W\times Z)$
		\For{each driven-frame cell centre $c$}
			\State $c'\leftarrow\mathrm{Rot}(\bar{\varphi})c+\bar{p}_{t+1}$ \Comment{ego frame $\to$ reference frame}
			\State $k'\leftarrow\lfloor(c'-b)/\ell\rfloor$ \Comment{$b$: lower grid corner}
			\State $X_{t+1}[c]\leftarrow X'_{t+1}[k']$ if $k'$ is in bounds; otherwise free
		\EndFor
	\end{algorithmic}
\end{algorithm}

\begin{algorithm}[h]
	\caption{Actor Erase-Print transition, with $X'_{t+1}$ the occupancy, $i$ the actor, $P$ the ego pose, $\bar{G}^{(i)}_{t+1}$ the actor pose in the ego frame, and $(w^{(i)},l^{(i)})$ the recorded box size}
	\label{alg:erase-print}
	\begin{algorithmic}[1]
		\State \textbf{Setup at selection frame} $t_0$
		\State \quad $pattern \leftarrow$ \Call{Capture}{actor body inside its recorded box at the GT pose}
		\State \quad \Call{Erase}{actor's own-class voxels inside the recorded box}
		\State \quad \Call{FillGround}{columns freed by the erase, from GT} \Comment{restore bottom ground}
		\State \textbf{Each transition} $t \rightarrow t{+}1$ while the actor is alive
		\For{fine step in the actor's rollout under its behavior}
			\If{\Call{OffRoad}{} or \Call{Obstacle}{} or \Call{EgoHit}{}}
				\State stop at the previous safe state \Comment{validation-before-paint}
			\EndIf
		\EndFor
		\State \Call{Stamp}{$pattern$, at reference-frame pose $P\circ\bar{G}^{(i)}_{t+1}$}
		\State After all actor edits, apply \Call{Recenter}{edited reference grid; $P$}
	\end{algorithmic}
\end{algorithm}

\subsection{Bounded Trajectory Deviation}
\label{sub:app3}
The following analysis applies to transitions whose driven ego position,
re-expressed in the reference-state frame (Eq.~\eqref{eq:warp}), satisfies
$\lVert\bar{p}_{t+1}\rVert_2\le\varepsilon$. This is a stated condition, not a
consequence of imitation learning or low open-loop ADE. The benchmark terminates
when driven-route deviation exceeds a threshold, following the use of off-route
termination in closed-loop evaluation \citep{bench2drive}. Because the check is
performed after a transition, its terminal state may overshoot the threshold;
the condition must be checked for each state to which the bound is applied.
The measured horizon scores are empirical and do not rely on this condition
holding for every evaluated frame.

\subsection{Bounded Operator Errors}
\label{sub:app4}
We compare each produced grid with its same-time recorded reference; this is
an alignment-and-edit difference, not an error against an observed
counterfactual scene. Let the cell size be $\ell$, the maximum planar cell-centre
range be $R$, and the ego pose deviations be at most $\varepsilon$ in position
and $\Delta\theta\in[0,\pi]$ in absolute heading. The inverse query of a centre
at range $r\le R$ moves by at most
\begin{equation}
 d_{\max}=\varepsilon+2R\sin(\Delta\theta/2)
 \le\varepsilon+R\Delta\theta.
 \label{eq:dmax}
\end{equation}
The factor $2\sin(\Delta\theta/2)$ is the rotation chord length; heading is in
radians. Extend the reference outside the grid with the free label. For class
$c$, let $C_c$ be its reference cells within the grid. In each fixed-height
slice, call a $c$-labelled cell a boundary cell if one of its four horizontal
face neighbors has a different label. Let $S_c$ count these cells within the
grid and its one-cell horizontal exterior halo, using the free-label extension.
Planar gathering leaves the height index unchanged. Set
$q=\lceil d_{\max}/\ell\rceil+1$ to conservatively cover voxel quantization.
For each output cell, the reference and sampled horizontal indices differ
by at most $q$ in either coordinate. If their class-$c$ indicators differ,
a horizontal face-neighbor path between them crosses a $c$ boundary; the
path lies within the index rectangle joining its endpoints. Hence the output
cell lies within horizontal index distance $q$ of a counted boundary cell at
the same height. The union of these squares contains at most
$S_c(2q+1)^2$ cells. This deliberately loose count requires no smoothness
assumption on the scene boundary.

Let $A_t$ count the output cells whose labels actor editing changes relative
to Ego Warp-Recenter alone at the same ego pose. It includes erasure, restored
ground, and stamping, and can affect background classes as well as actor classes.
With $V=HWZ$, the class symmetric difference $N_c=|C_c\mathbin{\triangle}C'_c|$
between the reference and final produced class set satisfies
\begin{equation}
 N_c\le \min\{V,\ S_c(2q+1)^2+A_t\}.
 \label{eq:nc-full}
\end{equation}
Here $A_t\le V$ is finite independently of an actor's accumulated route
separation; a per-frame speed cap alone does not bound that separation.
For $|C_c|>0$, write the false-positive and false-negative counts as
$F_c$ and $M_c$, so $N_c=F_c+M_c$ and the union has size $|C_c|+F_c$.
Therefore $1-\mathrm{IoU}_c=N_c/(|C_c|+F_c)\le N_c/|C_c|$, giving
\begin{equation}
 1-\mathrm{IoU}_c
 \le \min\!\left\{1,\frac{\min\{V,S_c(2q+1)^2+A_t\}}{|C_c|}\right\}.
 \label{eq:iou-error}
\end{equation}
This bound can be vacuous, particularly for thin or small classes, and is not a
guarantee of high counterfactual fidelity. The useful structural property is
that each occupancy grid is reconstructed from its own recorded reference,
rather than from a previous predicted grid. Thus occupancy-prediction errors
are not recursively fed back. Driven-pose deviation and actor edits can still
grow over time and affect fidelity, which is why the benchmark enforces a
valid operating range and reports empirical horizon scores.
\hfill $\square$

\subsection{Properties of Actor Erase-Print}
\label{sub:app5}
Controlled actor proposals are resolved before their bodies are painted.
The subsequent stamping is deterministic in the implementation's actor order.
It is order-independent only when rasterized supports are disjoint or write
the same label on their overlap; otherwise the last write wins. Erasure targets
the actor's own class, and stamping writes its class, but overwritten destination
voxels can previously belong to another class. A terminated actor is re-stamped
at its frozen pose after reference-frame updates, so it remains an obstacle
instead of vanishing. The captured pattern retains recorded actor geometry;
rotation, discretization, crop boundaries and overlapping stamps can alter its
rasterized support. Pattern reuse alone does not guarantee an in-distribution
tokenizer input. Obstacle and road checks operate with the implemented footprint
tests and tolerances; they are not a proof of zero static-voxel overwrite.
All resulting label changes, including restored ground, are covered by $A_t$
in Eq.~\eqref{eq:nc-full}.

\section{Simulator Details}
\label{app:simulator}

This appendix details the simulator, including the ego and actor kinematic models, the
controllers that execute plans at the substep cadence, the range-aware traffic
manager that selects which actors to re-drive, and the fidelity, baselines, and
tokenizer studies.

\subsection{Datasets}
\label{sub:app-datasets}
Table~\ref{tab:datasets} summarizes the occupancy datasets used in our
experiments.  All are ego-centred and metric.  Occ3D-nuScenes is recorded
natively at $2\,$Hz while Occ3D-Waymo is $10\,$Hz.  Occ3D-Waymo shares the
$0.4\,$m, $200{\times}200{\times}16$ ego-centred grid of Occ3D-nuScenes.

\begin{table}[h]
	\centering
	\caption{Dataset statistics of the occupancy benchmarks used, both from
	Occ3D \citep{occ3d}.}
	\footnotesize
	\begin{tabular}{@{}l l l@{}}
		\toprule
		\textbf{Property} & \textbf{Occ3D-nuScenes} &
		\textbf{Occ3D-Waymo} \\
		\midrule
		Source & nuScenes, validation & Waymo Open Dataset, validation \\
		Scenes / frames & 150 scenes, $6{,}019$ frames &
		202 scenes, ${\sim}39{,}987$ frames \\
		Frames per clip & ${\sim}40$ ($20\,$s) & ${\sim}198$ ($20\,$s) \\
		Native frequency & $2\,$Hz ($0.5\,$s) & $10\,$Hz ($0.1\,$s) \\
		Voxel resolution & $0.4\,$m & $0.4\,$m \\
		Point-cloud range & $[-40,-40,-1]$--$[40,40,5.4]\,$m &
		$[-40,-40,-1]$--$[40,40,5.4]\,$m \\
		Voxel grid & $200{\times}200{\times}16$ & $200{\times}200{\times}16$ \\
		Semantic classes & $18$ & $18$ \\
		\bottomrule
	\end{tabular}
	\label{tab:datasets}
\end{table}

\subsection{Substep Calibration}
\label{sub:app-substep}
The control executes $K$ integration substeps per occupancy frame
($\tau'=\tau/K$).  We fix the execution cadence at $20\,$Hz, giving
$K{=}20/\mathrm{fps}$, so $K{=}10$ at $2\,$Hz under the nuScenes/I$^2$-World protocol and
$K{=}2$ at the native $10\,$Hz.  \rev{Table~\ref{tab:substep-calib} sweeps $K$ over the
202-scene, $16\,$s protocol under GT replay.}  $K{=}2$ and $K{=}5$ are within noise
($92.84$ vs.\ $92.82$ IoU), $K{=}1$ and $K{=}3$ are lower, and $K{=}8$ collapses
to $63.0$ IoU, where the coarse interpolation of the six-waypoint reference
becomes unstable; we adopt the $20\,$Hz rule ($K{=}2$).

\begin{table}[h]
	\centering
	\caption{\rev{Substep calibration on the 202-scene, $16\,$s protocol.
	Values are IoU/mIoU; $K{=}2$ follows the $20\,$Hz rule.}}
	\footnotesize
	\begin{tabular}{@{}l ccccc@{}}
		\toprule
		\textbf{Dataset} & $K{=}1$ & $K{=}2$ & $K{=}3$ & $K{=}5$ & $K{=}8$ \\
		\midrule
		Occ3D-Waymo & \rev{0.892/0.863} & \rev{\textbf{0.928/0.916}} & \rev{0.925/0.912} &
		\rev{0.928/0.915} & \rev{0.630/0.489} \\
		\bottomrule
	\end{tabular}
	\label{tab:substep-calib}
\end{table}

\subsection{Fidelity Protocol}
\label{sub:app-persecond}

\begingroup\revisioncolor
Table~\ref{tab:persecond} lists the per-second values behind the Avg columns of
Tables~\ref{tab:fidelity} and~\ref{tab:waymo-sim}. At each reported
horizon, we pool voxel confusion matrices over the contributing episodes
and compute IoU/mIoU from the pooled counts. For the full-rollout Avg,
we instead pool over every evaluated frame:
\begin{equation}
C_{\mathrm{all}} = \sum_{i=1}^{N}\sum_{t=1}^{T_i} C_{i,t},
\qquad \mathrm{Avg} = \mathcal{M}(C_{\mathrm{all}}),
\label{eq:avg}
\end{equation}
where $C_{i,t}$ is a confusion matrix and $\mathcal{M}$ is the official
IoU or mIoU computation. Thus Avg is neither an arithmetic mean of the
16 horizon rows nor an average of per-frame IoUs. The contributing episode
count falls from 150 to 147 on nuScenes and from 190 to 164 on Waymo
(10\,Hz). The official mIoU implementation excludes the free class and
ignores zero-valued class IoUs when taking the class mean; we retain this
convention for consistency with the released evaluation code.
\endgroup

\begin{table}[h]
	\centering
	\revisioncolor
	\caption{\rev{Per-second occupancy fidelity on Occ3D-nuScenes (150 scenes,
	2\,Hz) and Occ3D-Waymo (202 scenes, 16\,s).  $n$ is the contributing
	episode count; \textbf{Avg} is computed from pooled confusion counts over the full rollout.}}
	\label{tab:persecond}
	\small
	\setlength{\tabcolsep}{5pt}%
	\begin{tabular}{@{}l ccc ccc ccc@{}}
		\toprule
		& \multicolumn{3}{c}{\textbf{nuScenes, 2\,Hz}} & \multicolumn{3}{c}{\textbf{Waymo, 10\,Hz}} & \multicolumn{3}{c}{\textbf{Waymo, 2\,Hz}}\\
		\cmidrule(r){2-4}\cmidrule(lr){5-7}\cmidrule(l){8-10}
		\textbf{Horizon} & \textbf{IoU} & \textbf{mIoU} & $n$ & \textbf{IoU} & \textbf{mIoU} & $n$ & \textbf{IoU} & \textbf{mIoU} & $n$\\
		\midrule
		$1\,$s & 75.13 & 74.92 & 150 & 93.18 & 93.57 & 190 & 93.56 & 95.09 & 191 \\
		$2\,$s & 76.10 & 74.48 & 150 & 92.29 & 90.99 & 187 & 93.61 & 94.98 & 189 \\
		$3\,$s & 75.84 & 74.82 & 150 & 92.28 & 90.03 & 183 & 94.20 & 94.50 & 184 \\
		$4\,$s & 76.63 & 78.27 & 150 & 92.42 & 90.38 & 179 & 94.42 & 95.94 & 180 \\
		$5\,$s & 79.29 & 79.86 & 150 & 92.78 & 92.39 & 176 & 95.93 & 96.60 & 177 \\
		$6\,$s & 78.92 & 77.31 & 149 & 93.42 & 92.94 & 175 & 96.37 & 96.33 & 176 \\
		$7\,$s & 77.76 & 74.68 & 149 & 94.21 & 93.59 & 174 & 95.99 & 96.05 & 176 \\
		$8\,$s & 75.43 & 72.60 & 147 & 93.53 & 92.68 & 172 & 95.11 & 96.02 & 174 \\
		$9\,$s & 74.03 & 70.28 & 147 & 93.24 & 92.58 & 172 & 95.19 & 95.45 & 173 \\
		$10\,$s & 74.03 & 73.86 & 147 & 92.42 & 90.58 & 172 & 96.47 & 97.08 & 172 \\
		$11\,$s & 74.80 & 73.39 & 147 & 92.68 & 91.48 & 171 & 96.31 & 95.53 & 172 \\
		$12\,$s & 75.34 & 73.06 & 147 & 92.97 & 90.99 & 169 & 95.60 & 93.67 & 170 \\
		$13\,$s & 73.18 & 70.81 & 147 & 92.86 & 91.35 & 169 & 96.30 & 94.48 & 170 \\
		$14\,$s & 75.79 & 73.55 & 147 & 91.68 & 89.22 & 168 & 96.50 & 95.04 & 170 \\
		$15\,$s & 74.72 & 73.63 & 147 & 92.61 & 91.45 & 167 & 96.94 & 96.29 & 170 \\
		$16\,$s & 74.49 & 73.93 & 147 & 92.83 & 92.83 & 164 & 94.96 & 92.33 & 167 \\
		\midrule
		\textbf{Avg} & \textbf{75.6} & \textbf{74.1} & -- & \textbf{92.84} & \textbf{91.65} & -- & \textbf{95.44} & \textbf{95.29} & --\\
		\bottomrule
	\end{tabular}
\end{table}

\subsection{Teacher-Forcing Protocols}
\label{sub:app15}
We compare the learned baselines at their strongest, one-frame-lookahead
setting, which isolates the effect of the transition operators.  At each step the
baseline is fed the \emph{recorded} reference frame, so no error compounds and
the reference frame is kept largely unaffected; any remaining gap is therefore
attributable to the fidelity of the learned transition rather than to error
accumulation.  For I$^2$-World \citep{i2world}, we run the released checkpoint
under per-step teacher-forcing, feeding back the \emph{recorded} frame token and
ego motion at each step, with the released evaluation settings (same
Occ3D-nuScenes windows, ego conditioning, frozen tokenizer, and official
\emph{forecasting\_miou}).  The I$^2$-World-TF rows therefore measure
reconstruction-level fidelity given perfect per-step context, not a multi-step
forecast.  For DOME \citep{dome}, the DOME-TF row applies the same protocol with
the released diffusion model, which generates a block of future occupancy from a
recorded context window.  To match the one-frame-lookahead comparison, we
condition on the \emph{recorded} window of up to four context frames ending at
the current frame, generate the block, and score only its \emph{first} frame
against the recorded next frame.  The rest is discarded and no generated frame is
fed back, so DOME enters at its strongest in-distribution point.  The
long-horizon curve aggregates this lookahead over the same 150 val scenes as
WorldOcc.  Table~\ref{tab:fidelity-tf} reports both.  Even with this oracle
context, both patched baselines still fall short of WorldOcc, showing the limited
fidelity of their learned models.

\begin{table}[h]
	\centering
	\caption{Long-horizon occupancy fidelity in IoU/mIoU \rup (\%) under
	per-step teacher-forcing.  \rev{\textbf{Avg}: reported-horizon mean
	(baselines), whole-horizon aggregate (WorldOcc).}}
	\label{tab:fidelity-tf}
	\footnotesize
	\setlength{\tabcolsep}{3.5pt}%
	\begin{tabular}{@{}l l ccccc c@{}}
		\toprule
		\textbf{Method} & \textbf{Year} & 1\,s & 2\,s & 3\,s & 8\,s & 16\,s & \textbf{Avg}\\
		\midrule
		DOME-TF \citep{dome} & 2024 & 47.6/41.8 & 49.6/43.0 & 49.0/42.4 & 48.7/42.3 & 49.1/41.2 & 48.8/42.1\\
		I$^2$-World-TF \citep{i2world} & 2025 & 56.6/53.8 & 57.2/56.1 & 57.2/55.3 & 58.3/56.5 & 57.4/54.7 & 57.4/55.3\\
		\rowcolor{rowlight}\textbf{WorldOcc} (ours) & 2026 & 75.1/74.9 & 76.1/74.5 & 75.8/74.8 & 75.4/72.6 & 74.5/73.9 & \textbf{\rev{75.6/74.1}}\\
		\bottomrule
	\end{tabular}
\end{table}

\subsection{Perceptual Fidelity}
\label{sub:app-perceptual}
We additionally compare the realism of the generated occupancy itself with the
generative-metric protocol of SemCity~\citep{semcity}.  Following SemCity's
renderer, every frame is rendered to a $2$D image with \texttt{pyrender} using
the front camera, and scored with default \texttt{torch-fidelity} InceptionV3 to
report the Inception Score (IS) on the generated frames, and the Fr\'echet
Inception Distance (FID) and Kernel Inception Distance (KID) between the
generated and reference distributions.  All methods are
evaluated on the same Occ3D-nuScenes val windows at the
matched $1/2/3\,$s horizon ($2/4/6$ future frames at $2\,$Hz), with the recorded
GT target frames of those windows, rendered identically, as the reference
distribution.  DynamicCity \citep{dynamiccity} is omitted as its conditional
model weight on this dataset is not released.  Table~\ref{tab:perceptual-nuscenes} reports
that WorldOcc attains the best FID and KID.  Its rigid-operator transition preserves the
recorded scene by construction, so its re-centred frames stay close to the
reference, whereas the learned generators trade fidelity for diversity.

\begin{table}[h]
	\centering
	\caption{Perceptual metrics on Occ3D-nuScenes at the matched $1/2/3\,$s
	horizon.  DynamicCity \citep{dynamiccity} is omitted as its conditional
	model weight on this dataset is not released.}
	\footnotesize
	\begin{tabular}{@{}l l ccc@{}}
		\toprule
		\textbf{Method} & \textbf{Year} & IS \rup & FID \rdown & KID \rdown \\
		\midrule
		DOME \citep{dome} & 2024 & 2.046 & 13.13 & 0.0081 \\
		OccWorld \citep{occworld} & 2024 & 2.173 & 24.47 & 0.0187 \\
		I$^2$-World \citep{i2world} & 2025 & 1.925 & 36.95 & 0.0287 \\
		\rowcolor{rowlight}\textbf{WorldOcc} (ours) & 2026 & \textbf{2.210} & \textbf{1.06} & \textbf{0.00004} \\
		\bottomrule
	\end{tabular}
	\label{tab:perceptual-nuscenes}
\end{table}

\subsection{Agent States}
\label{sub:app6}
The ego state is $E_t=(x_t,y_t,\theta_t,v_t,\delta_t,\alpha_t,g_t)$ (pose
$(x,y,\theta)$, speed $v$, steering $\delta$, acceleration $\alpha$, and a
one-hot navigation command $g_t\in\{0,1\}^3$).  Each actor's state is
$G^{(i)}_t=(x^{(i)}_t,y^{(i)}_t,\theta^{(i)}_t,v^{(i)}_t,c^{(i)},w^{(i)},l^{(i)})$,
carrying pose and speed with class $c^{(i)}$ and box size $(w^{(i)},l^{(i)})$.
Only the driven poses $E_{t+1},G_{t+1}$ enter the transition $T$
(Eq.~\eqref{eq:transition-ref}).  The command and each actor's class and size are
static attributes consumed by the planners, operators, and traffic manager.  In
particular, the recorded box $(G'^{(i)}_{t_0},w^{(i)},l^{(i)})$ defines the
footprint of Actor Erase-Print, and capture, erasure, and ground restoration are all
restricted to it.

\subsection{Bicycle Ego Kinematic}
\label{sub:app7}
The ego is a kinematic bicycle (wheelbase $L$, frame period $\tau$, state
$E=(x,y,\theta,v,\delta,\alpha)$).  Its time derivative
$\dot E=(\dot x,\dot y,\dot\theta,\dot v,\dot\delta,\dot\alpha)$, i.e.\
$\dot E=\mathrm{d}E/\mathrm{d}t$, is
\begin{equation}
	\dot E=\Big(v\cos\theta,\;v\sin\theta,\;\tfrac{v\tan\delta}{L},\;\alpha,\;\dot\delta,\;\dot\alpha\Big),
	\label{eq:bicycle}
\end{equation}
which, discretized with the fine Euler step $\tau'=\tau/K$ at $K$ substeps per
frame, gives the update
\begin{equation}
	v' = v+\alpha\,\tau', \quad
	\delta' = \delta+\dot\delta\,\tau', \quad
	\theta' = \theta+\frac{v'\tan\delta'}{L}\,\tau', \quad
	p' = p+v'\begin{bmatrix}\cos\theta\\ \sin\theta\end{bmatrix}\tau',
	\label{eq:bstep}
\end{equation}
where $\alpha$ and $\dot\delta$ are the applied acceleration and steering rate.
A single coarse control hold per frame makes the driven pose oscillate on a
curved route, so we separate the occupancy frame rate from a fixed $20\,$Hz
execution rate by adjusting the substeps $K$.  This keeps the driven pose close
to the recording.
Raising $K$ from $1$ to $10$ cuts the driven-vs-GT error from
$\sim$0.9\,m/$6.5\degree$ to $\sim$0.1\,m/$0.4\degree$, and the trend plateaus
beyond $K{=}10$.  It refines only the low-level execution of the emitted plan and
leaves the route unchanged.

At each frame the planner emits six waypoints $0.5\,$s apart.  iLQR cubic-splines
them into $6K$ fine waypoints and solves one full-horizon problem at the fine
step $\tau'=\tau/K$, producing $K$ control samples that are applied one per
substep.  It therefore solves once per frame, not per substep, and adds no
waypoints beyond the spline.  The final substep expresses the driven pose in the next GT frame,
as Ego Warp-Recenter expects.  Prior simulators (e.g.\ HUGSIM \citep{hugsim})
apply a single coarse iLQR hold per frame, which under-suppresses the
high-frequency curvature present in the noisy recorded trajectory.

\subsection{Unicycle Actor Kinematic}
\label{sub:app8}
Each actor is a unicycle with state $G^{(i)}=(x^i,y^i,\theta^i,v^i)$, where
$p^i=(x^i,y^i)$ is the actor's position, and control $(\alpha^i,\omega^i)$.
Its time derivative
$\dot G^{(i)}=(\dot x^i,\dot y^i,\dot\theta^i,\dot v^i)$, i.e.\
$\dot G^{(i)}=\mathrm{d}G^{(i)}/\mathrm{d}t$, is
\begin{equation}
	\dot G^{(i)}=\Big(v^i\cos\theta^i,\;v^i\sin\theta^i,\;\omega^i,\;\alpha^i\Big),
	\label{eq:uni}
\end{equation}
with per-class limits on speed, acceleration, deceleration and yaw rate, and
discretized as
\begin{equation}
	v' = \max(0,\,v+\alpha\,\tau'), \quad
	\theta' = \theta+\omega\,\tau', \quad
	p' = p+v'\begin{bmatrix}\cos\theta'\\ \sin\theta'\end{bmatrix}\tau'.
	\label{eq:ustep}
\end{equation}
The action $a^{(i)}_t\in\mathbb{R}^2$ is an \emph{aim point} specifying a target
offset from the actor.  The controller steers toward it with a
pure-pursuit law,
\begin{equation}
	e_\theta=\operatorname{atan2}\big(w^{i}_{\text{aim}}-p\big)-\theta,
	\qquad
	\omega=\operatorname{clamp}\big(k\,e_\theta,\pm\omega_{\max}\big),
	\label{eq:aim}
\end{equation}
where $\theta$ is the actor's heading, $e_\theta$ the bearing to the aim point,
$k$ the proportional gain, and $\omega_{\max}$ the per-class yaw-rate limit.
The desired speed $v^{*}$ follows a braking-distance profile that
leaves the actor able to stop before the aim point, clamped by the per-class
maximum speed $v_{\max}(c)$,
\begin{equation}
	v^{*}=\min\Big(v_{\max},
	\sqrt{2\,a_{\text{dec}}\,\|w^{i}_{\text{aim}}-p\|}\Big)\,
	\max\big(0,\cos e_\theta\big),
	\label{eq:speed}
\end{equation}
where $a_{\text{dec}}$ is the per-class deceleration limit.  The per-class
speed cap $v_{\max}(c)$ decodes the per-frame displacement cap $d_c(c)$ at the
control period $\Delta t$,
\begin{equation}
	v_{\max}(c) = \frac{d_c(c)}{\Delta t},
	\label{eq:speed-cap}
\end{equation}
A per-substep safety brake stops the actor before overlapping a
non-traversable cell or the ego, holding one substep short of contact.

\subsection{The Controllers}
\label{sub:app9}
Following \citet{nuplan}, the ego's iterative Linear Quadratic Regulator (iLQR)
controller interpolates the six emitted waypoints with a cubic spline into
$6K$ fine waypoints, then solves a receding-horizon tracking problem whose
control sequence is re-solved at the frame cadence and consumed at the fine
substep rate.  Each selected actor is driven by an IDM-inspired
waypoint-following controller that tracks the actor's emitted aim point with
the per-class kinematic clamps of Eq.~\eqref{eq:speed-cap} and an obstacle
safety brake.  It follows IDM's spirit of smooth acceleration toward a desired
speed and safe braking, but not IDM's car-following formula, since it follows
an arbitrary waypoint rather than a lead vehicle \citep{idm}.

\begin{algorithm}[h]
	\caption{Substep iLQR ego control, executed once per frame.  $E_t$ is the ego state, $\{w_j\}_{j=1}^{6}$ the six emitted waypoints, $K$ the substeps per frame, and $\tau'=\tau/K$ the fine step.}
	\label{alg:ilqr}
	\begin{algorithmic}[1]
		\State $r \leftarrow$ cubic-spline$\big(\{w_j\}\big)$ sampled at $\tau'$ \Comment{$6K$ fine waypoints}
		\State $u \leftarrow$ \text{iLQR}($E_t, r$) \Comment{standard iLQR \citep{nuplan}, with a warm start followed by forward rollout, backward Riccati pass, and forward pass to convergence}
		\For{$j=0,\dots,K{-}1$} \Comment{one fine step $\tau'=\tau/K$ per substep}
			\State $\alpha_j, \dot\delta_j \leftarrow u_j$ \Comment{applied acceleration and steering rate}
			\State $v \leftarrow v+\alpha_j\,\tau'$ \Comment{Eq.~\eqref{eq:bstep}}
			\State $\delta \leftarrow \delta+\dot\delta_j\,\tau'$
			\State $\theta \leftarrow \theta+\frac{v\tan\delta}{L}\,\tau'$
			\State $p \leftarrow p+v\begin{bmatrix}\cos\theta\\ \sin\theta\end{bmatrix}\tau'$
		\EndFor
	\end{algorithmic}
\end{algorithm}

\subsection{Traffic Manager}
\label{sub:app10}
Each actor class $c$ has a per-frame displacement cap $d_c(c)$, so an actor
within the reachable radius $F\,d_c(c)$ can reach the ego within $F=6$ frames.
For temporal diversity we draw the start
frame $t_0\sim\mathrm{Uniform}(\{0,\dots,\min(F,M)\})$, with $M$ the number of
frames remaining in the clip, walk the frames $t=t_0,\dots,\min(F,M)$, and at
each construct the candidate set $\mathcal{C}_t$ of actors that are not yet
selected, are valid at $t$, and lie within their class reach.  With
$\mathcal{P}$ the participant set, $\mathcal{R}$ the re-driven set, $p_i(t)$
participant $i$'s ego-centric pose, and $m_i(t)$ its validity mask,
\begin{equation}
	\mathcal{C}_t = \bigl\{\, i \in \mathcal{P} \setminus \mathcal{R} :
	m_i(t) = 1,\ \ \lVert p_i(t)\rVert_2 \le F\,d_c\bigl(c(i)\bigr) \,\bigr\},
	\label{eq:cand}
\end{equation}
At the first frame with a non-empty candidate set, we draw one actor uniformly,
$i^{\star}\sim\mathrm{Uniform}(\mathcal{C}_t)$, update
$\mathcal{R}\leftarrow\mathcal{R}\cup\{i^{\star}\}$, and continue until $k$
actors are drawn or the frames run out.

\begin{figure}[h]
	\centering
	\makebox[\cellw]{Straight}\hspace{\cellgap}%
	\makebox[\cellw]{Turn left}\hspace{\cellgap}%
	\makebox[\cellw]{Turn right}\\[1pt]
	\displabel{(+5.60\,m, -0.03\,m)\ (+10.81\,m, -0.05\,m)\ (+15.66\,m, -0.08\,m)}%
	{(+3.71\,m, +0.16\,m)\ (+6.55\,m, +0.99\,m)\ (+8.42\,m, +2.24\,m)}%
	{(+3.80\,m, -0.17\,m)\ (+6.72\,m, -1.05\,m)\ (+8.46\,m, -2.53\,m)}\\[1pt]
	\subcaptionbox{WorldOcc}{%
		\includegraphics[height=\cellh]{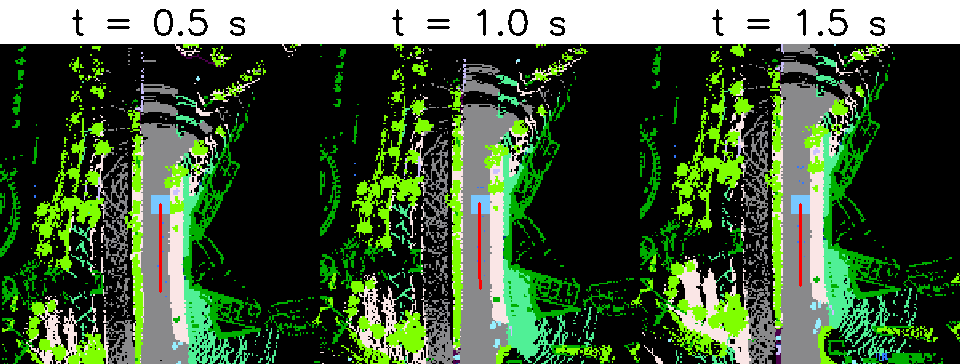}\hspace{\cellgap}%
		\includegraphics[height=\cellh]{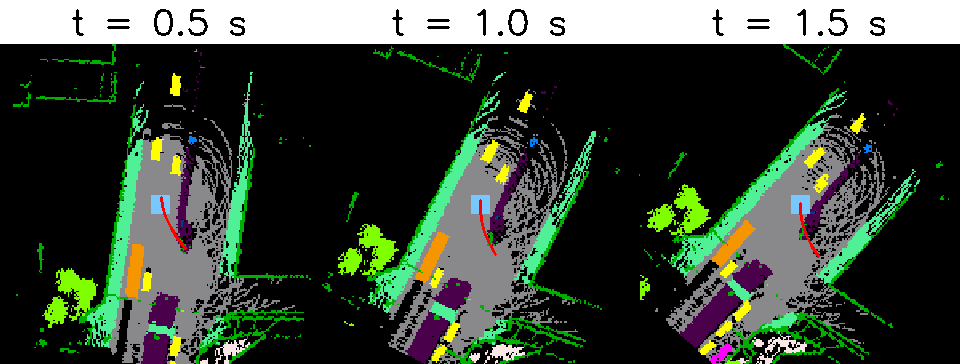}\hspace{\cellgap}%
		\includegraphics[height=\cellh]{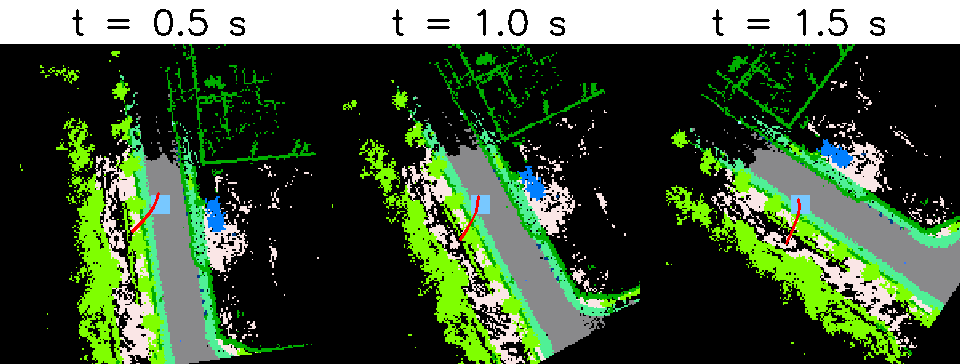}}\\[1pt]
	\subcaptionbox{I$^2$-World}{%
		\includegraphics[height=\cellh]{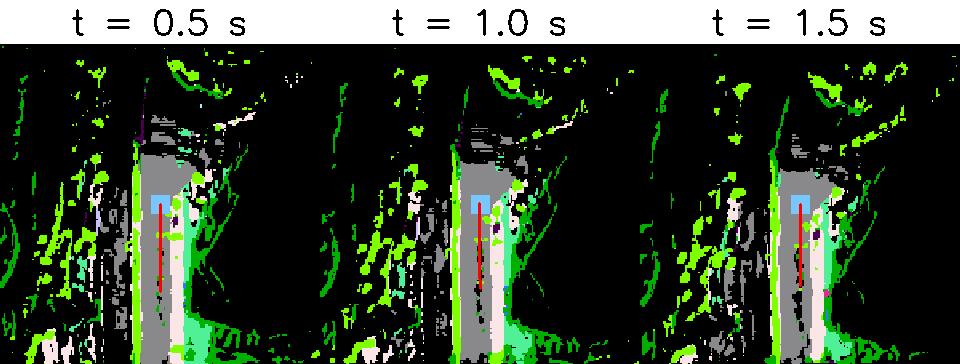}\hspace{\cellgap}%
		\includegraphics[height=\cellh]{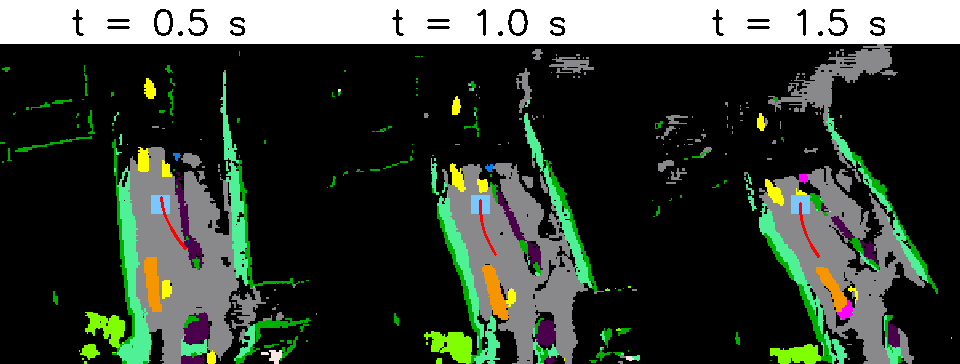}\hspace{\cellgap}%
		\includegraphics[height=\cellh]{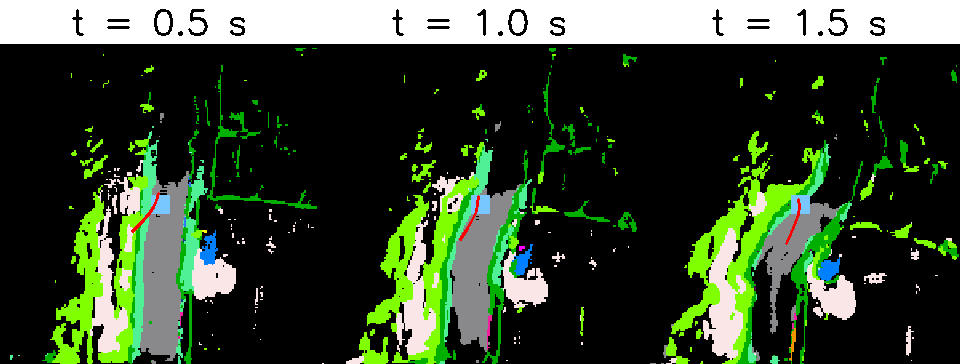}}\\[1pt]
	\subcaptionbox{DOME}{%
		\includegraphics[height=\cellh]{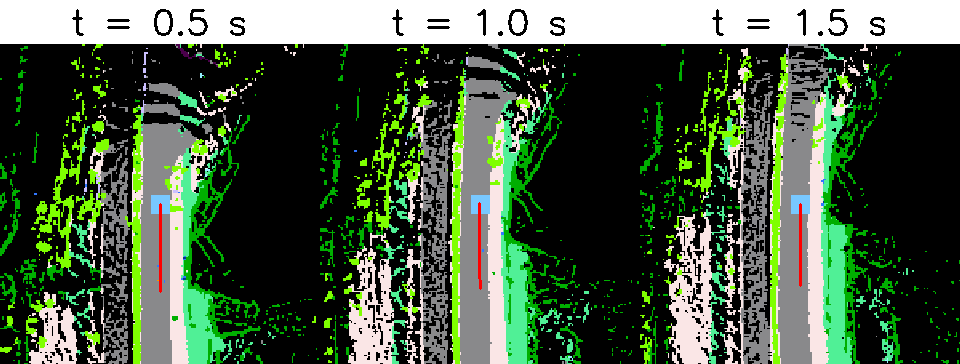}\hspace{\cellgap}%
		\includegraphics[height=\cellh]{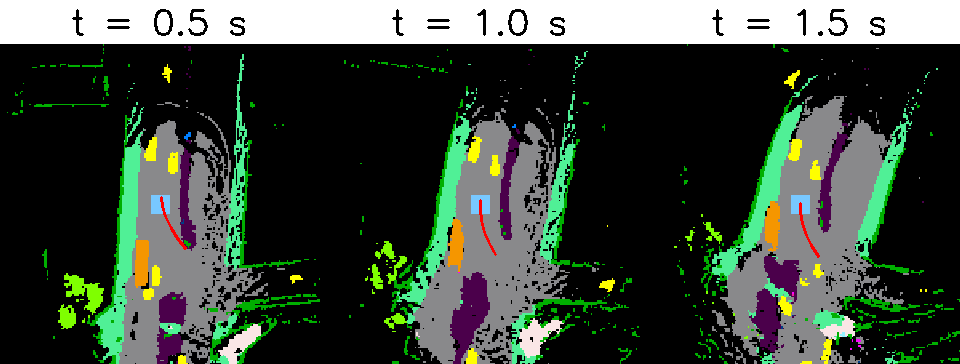}\hspace{\cellgap}%
		\includegraphics[height=\cellh]{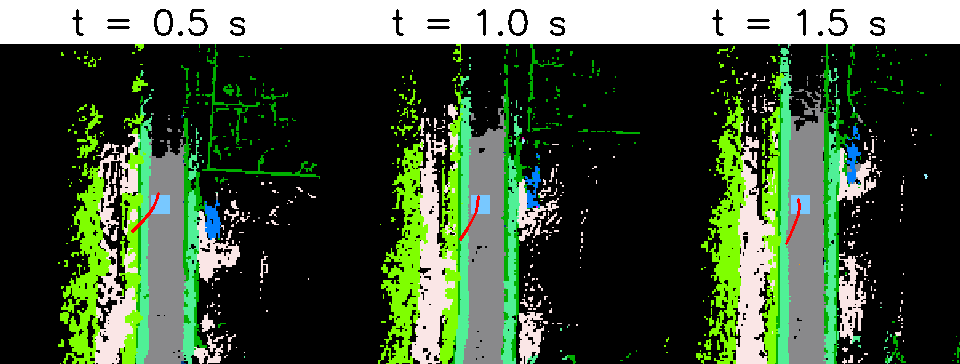}}
	\caption{Controllability under a counterfactual natural turning command.}
	\label{fig:control}
\end{figure}

\subsection{Operator Timing}
\label{sub:app13}
The main-text throughput is the operator-only rate on an RTX\,4090, excluding the
controller and perception.  Ego Warp-Recenter takes $0.79$\,ms, about $1270$\,FPS, and Actor Erase-Print
$3.58$\,ms, about $280$\,FPS.  On CPU with numba they take
$2.3$\,ms, about $436$\,FPS, and $12.4$\,ms, about $81$\,FPS.

\subsection{Sensitivity of Occupancy IoU to the Driven Pose Deviation}
\label{sub:app17}
Under a controlled pose deviation, IoU is highly sensitive to heading error
($0.5\degree\to68\%$, $1\degree\to50\%$) and far less to translation, as
sub-15\,cm offsets are absorbed by the $0.4$\,m voxels and it falls only from
20\,cm (Table~\ref{tab:deviation-sensitivity}).  Given the measured tracking deviations (see the controlled perturbation
results), much of the residual IoU loss is the metric's rotation sensitivity at
the controller's tracking error, not scene degradation.  At this heading floor
the reported 70--80\% IoU is near the metric's ceiling, which WorldOcc holds flat
while the learned baselines decay.

\begin{table}[h]
	\centering
	\caption{Occupancy IoU / mIoU under a controlled driven-vs-GT deviation,
	using pure translation (cm) or pure heading rotation (deg).}
	\footnotesize
	\subcaptionbox{Translation (cm)}{%
	\begin{tabular}{@{}l ccccccccc@{}}
		\toprule
		\textbf{dev (cm)} & 0 & 5 & 10 & 15 & 20 & 30 & 40 & 60 & 80\\
		\midrule
		\textbf{IoU} & 1.000 & 1.000 & 1.000 & 1.000 & 0.628 & 0.546 & 0.546 & 0.448 & 0.415\\
		\textbf{mIoU} & 1.000 & 1.000 & 1.000 & 1.000 & 0.597 & 0.517 & 0.517 & 0.392 & 0.350\\
		\bottomrule
	\end{tabular}}\\[2pt]
	\subcaptionbox{Heading (deg)}{%
	\begin{tabular}{@{}l cccccccccc@{}}
		\toprule
		\textbf{dev (deg)} & 0.0 & 0.5 & 1.0 & 1.5 & 2.0 & 3.0 & 4.0 & 5.0 & 6.0 & 8.0\\
		\midrule
		\textbf{IoU} & 1.000 & 0.681 & 0.495 & 0.416 & 0.366 & 0.304 & 0.266 & 0.240 & 0.220 & 0.193\\
		\textbf{mIoU} & 1.000 & 0.694 & 0.486 & 0.376 & 0.304 & 0.221 & 0.173 & 0.140 & 0.118 & 0.089\\
		\bottomrule
	\end{tabular}}
	\label{tab:deviation-sensitivity}
\end{table}

\begingroup\revisioncolor
While IoU is the de facto standard for occupancy evaluation and we report it for
comparability, it is a known pose-sensitive metric: a global heading or
translation offset shifts far-range cells by more than one voxel, so the score
conflates scene fidelity with ego-pose alignment.  This effect is strongly range-dependent
(Table~\ref{tab:range-sensitivity}): at $0.5\degree$ the near field is
essentially unchanged while the far field already drops to about $0.57$.  This is a limitation of the metric and of the field rather than of the
simulator alone: a pose offset changes the coordinate sampling, while discretization
and finite-field cropping also change the produced grid. Appendix~\ref{sub:app4}
gives a conditional reference-relative bound, not a guarantee that error stays
small over a rollout. Occupancy predictions are not recursively fed back, but
driven poses and actor edits can diverge over time. We
therefore read the reported IoU as a conservative, pose-sensitive fidelity
measure, and consider a pose-robust occupancy-fidelity metric an important
direction for the field.

\begin{table}[h]
	\centering
	\revisioncolor
	\caption{\rev{Occupancy IoU under small pose perturbations, averaged over 20
	distinct validation scenes (one first-frame occupancy per scene),
	split by radial range. Near field: $<20$\,m; far field: $\ge20$\,m.}}
	\label{tab:range-sensitivity}
	\footnotesize
	\begin{tabular}{@{}l ccc@{}}
		\toprule
		\textbf{Perturbation} & \textbf{All} & \textbf{Near} ($<20$\,m) & \textbf{Far} ($\ge20$\,m)\\
		\midrule
		Rotation $0.5\degree$ & 0.702 & 1.000 & 0.567 \\
		Rotation $1.0\degree$ & 0.510 & 0.769 & 0.390 \\
		Rotation $2.0\degree$ & 0.374 & 0.605 & 0.267 \\
		Translation $0.2$\,m & 0.633 & 0.734 & 0.581 \\
		Translation $0.4$\,m & 0.558 & 0.669 & 0.501 \\
		\bottomrule
	\end{tabular}
\end{table}

\begin{figure}[h]
	\centering
	\includegraphics[width=0.8\textwidth]{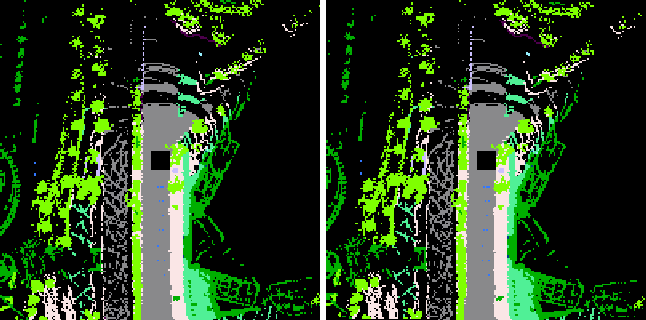}
	\revisioncolor
	\caption{\rev{Same scene at $0\degree$ (left) and $1\degree$ (right) heading
	deviation.  The two renders are visually near-identical, while IoU drops from
	$1.00$ to $0.43$ for this scene ($0.50$ averaged over scenes), showing that
	voxel IoU is rotation-sensitive rather than the scene being degraded.}}
	\label{fig:heading-sensitivity}
\end{figure}
\endgroup

\subsection{Deviation Threshold}
\label{sub:app20}
The deviation threshold $d_{\text{dev}}$ is applied \emph{after} Ego
Warp-Recenter, so it also bounds how much of the finite $80\times80$\,m scene can
leave the window.  Both frames are $80\times80$\,m, so after re-centring at a
pose displaced by $d$ the retained scene is the overlap of two co-sized squares.
For translation alone with no heading change, and the $1$--$15$\,m
range tabulated below, the minimum overlap is on the diagonal, offset $(d/\sqrt2,d/\sqrt2)$, giving
\begin{equation}
	\rho_{\text{ret}}(d)=\Bigl(\tfrac{80-d/\sqrt{2}}{80}\Bigr)^{2},
	\label{eq:dev-retention}
\end{equation}
a corner band of width $d/\sqrt2$.  For example, $5$\,m keeps at least $91\%$ of
the scene under this translation-only calculation and $10$\,m about $83\%$ (Table~\ref{tab:dev-retention}), so we adopt
$d_{\text{dev}}=5$\,m, which leaves
planners enough range to react to varying traffic while retaining a valid
fraction of the in-bound scene. Rotation can reduce this overlap further; the
translation-only calculation is not a joint position-and-heading guarantee.

\begin{table}[h]
	\centering
	\caption{Retained scene fraction $\rho_{\text{ret}}(d)=((80-d/\sqrt2)/80)^2$
	under translation $d$ on the worst-case diagonal for the tabulated range,
with no heading change.}
	\footnotesize
	\setlength{\tabcolsep}{4pt}%
	\begin{tabular}{@{}l ccccccccccccccc@{}}
		\toprule
		$d$ (m) & 1 & 2 & 3 & 4 & 5 & 6 & 7 & 8 & 9 & 10 & 11 & 12 & 13 & 14 & 15\\
		\midrule
		$R$ (\%) & 98 & 97 & 95 & 93 & 91 & 90 & 88 & 86 & 85 & 83 & 82 & 80 & 78 & 77 & 75\\
		\bottomrule
	\end{tabular}
	\label{tab:dev-retention}
\end{table}

\section{Actor Policy Details}
\label{app:actor}
\label{sub:app21}

\subsection{Scripted Baselines}
\label{sub:app22}
A re-driven actor is controlled from its start frame until the end of the clip
and follows one of two scripted behaviors.  \emph{Normal scripted}
replays the actor's recorded trajectory by aiming at its next GT waypoint, and
\emph{aggressive scripted} aims at the ego's current position, the origin of the
current ego-centric frame, with no future or ground-truth information.  Each
behavior defines an aim target
\begin{equation}
	w^{i}_{\text{aim}} = \begin{cases}
		\text{next GT waypoint}, & \text{normal scripted},\\
		\mathbf{0}, & \text{aggressive scripted},
	\end{cases}
	\label{eq:aim-idm}
\end{equation}
and the controller of Appendix~\ref{app:simulator} steers the actor's heading toward
this target while following the braking-distance speed profile of
Eq.~\eqref{eq:speed}.  Actors that are not re-driven keep their recorded motion.
The learned policies below share this aim-target and controller interface,
emitting a normalized action scaled by $d_c(c)$ into an aim offset.

\subsection{Naturalness Metrics}
\label{sub:app23}
\begin{figure}[h]
	\centering
	\includegraphics[width=0.9\textwidth]{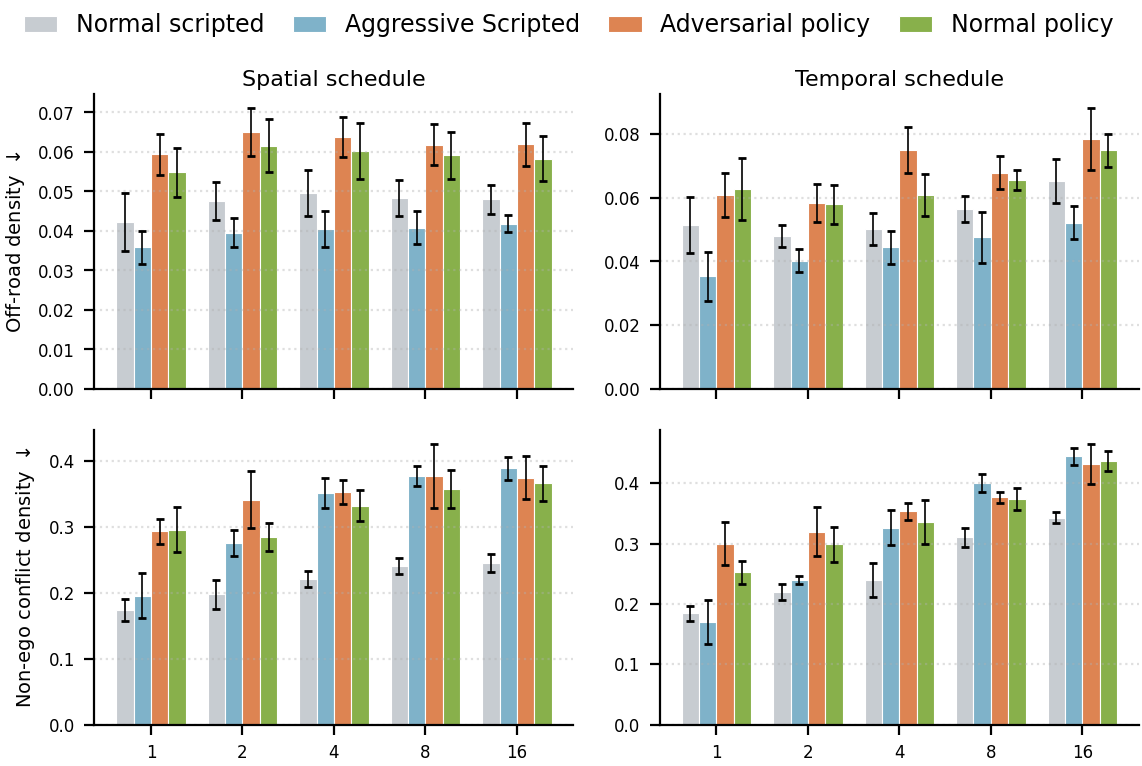}
	\caption{Off-road and non-ego conflict densities over the spatial and temporal
	$k$-sweeps, averaged over five seeds with std error bars.  Ego collision rate is in
	Fig.~\ref{fig:actor-scaling}.}
	\label{fig:actor-3metrics}
\end{figure}

To complement the ego collision rate (Fig.~\ref{fig:actor-scaling}), we report
two naturalness metrics for the controlled actors.  \emph{Off-road density} is
the fraction of controlled-actor frames whose footprint is majority off the
drivable area, and \emph{non-ego conflict density} the fraction on which the
safety brake fires against a non-ego object (Fig.~\ref{fig:actor-3metrics}).

\begin{figure}[h]
	\centering
	\begin{tikzpicture}
	\node[anchor=south west,inner sep=0] (img) at (0,0)
		{\includegraphics[width=0.86\textwidth]{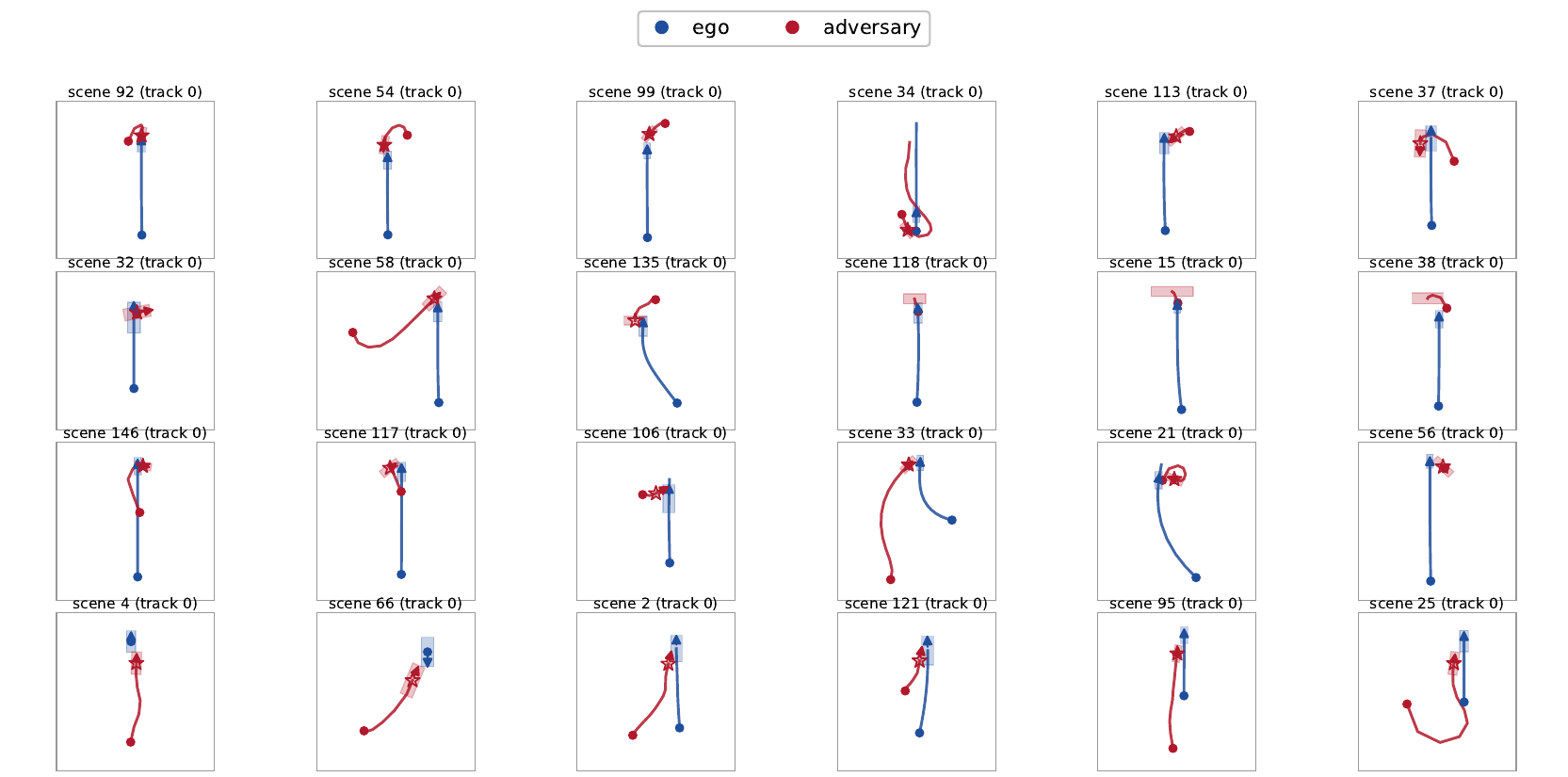}};
	\node[rotate=90,anchor=center,align=center,font=\scriptsize\bfseries] at ($(img.south west)!0.125!(img.north west)+(-0.14in,0)$) {Rear-end\\ (ego leads)};
	\node[rotate=90,anchor=center,align=center,font=\scriptsize\bfseries] at ($(img.south west)!0.375!(img.north west)+(-0.14in,0)$) {Crossing\\ paths};
	\node[rotate=90,anchor=center,align=center,font=\scriptsize\bfseries] at ($(img.south west)!0.625!(img.north west)+(-0.14in,0)$) {Rear-end\\ (ego follows)};
	\node[rotate=90,anchor=center,align=center,font=\scriptsize\bfseries] at ($(img.south west)!0.875!(img.north west)+(-0.14in,0)$) {Opposite\\ direction};
\end{tikzpicture}
	\revisioncolor
	\caption{\rev{Representative conflicts of the learned adversary at closest
	approach, one row per family of Fig.~\ref{fig:conflict-bars}.  Ego is blue
	(boxed, pointing up), adversaries red; dots mark the first frame, stars
	collisions.}}
	\label{fig:conflict-examples}
\end{figure}
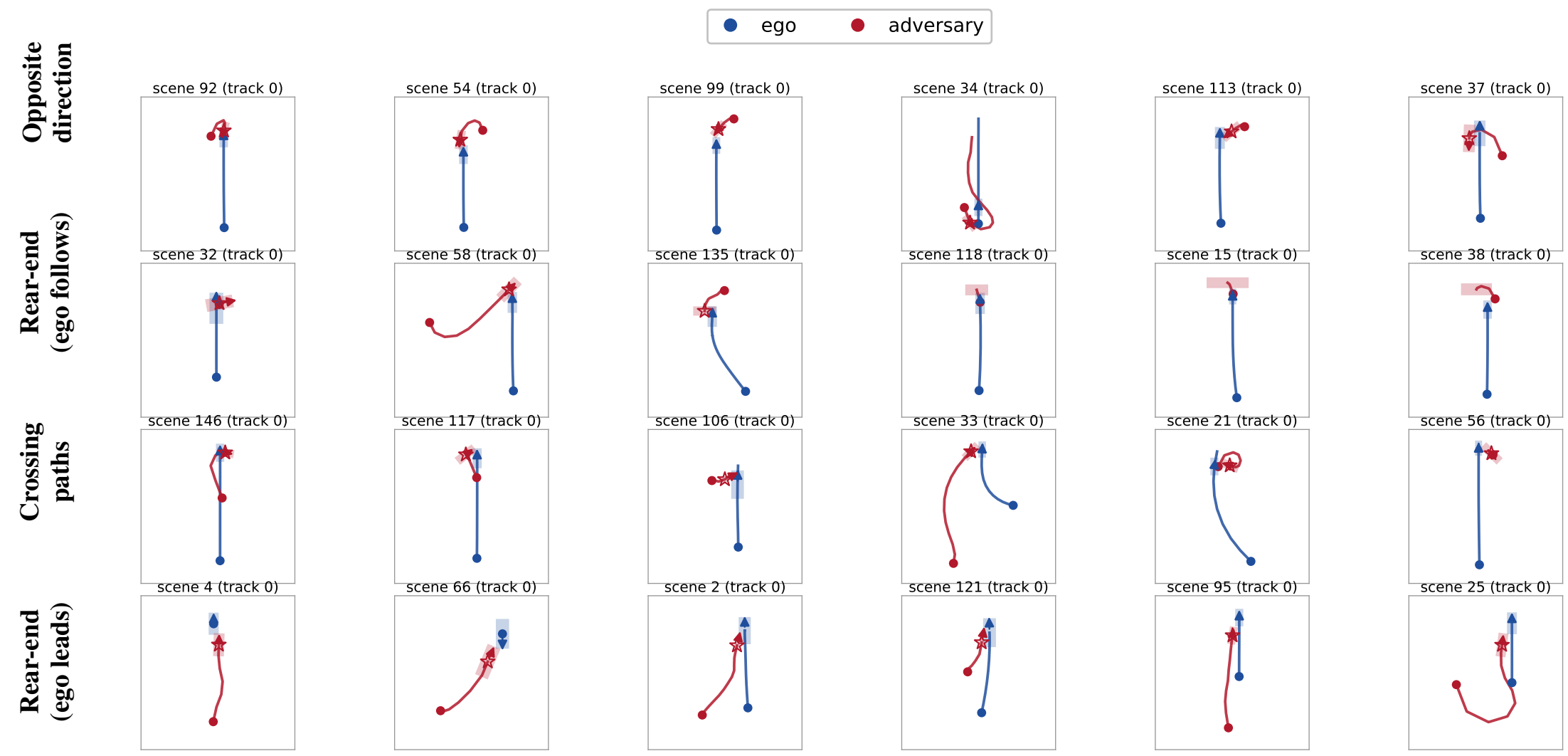

\subsection{Adversarial Traffic Behavior}
\label{sub:app24}
The adversarial actor is a shared-weight policy trained with single-agent PPO
in an independent-learning (IPPO-style) setup \citep{schroederdewitt2020ippo}
and deployed independently across actors.  One network reads each
controlled actor's localized scene context and drives the whole fleet.  The
objective is an expectation over single-actor episodes, each actor maximizing
the standard PPO clipped surrogate over its own reward $r^{(i)}_t$
(Eq.~\eqref{eq:actor-reward}),
\begin{equation}
	\mathcal{L}^{\mathrm{CLIP}}(\theta)
=\mathbb{E}_{i,t}\Big[\min\big(
\rho^{(i)}_t(\theta)\,\hat{A}^{(i)}_t,\,
\operatorname{clip}\!\big(\rho^{(i)}_t(\theta),\,1-\epsilon,\,1+\epsilon\big)
\,\hat{A}^{(i)}_t\big)\Big],
\label{eq:ppo-obj}
\end{equation}
where $\rho^{(i)}_t$ is the importance-sampling ratio, $\hat{A}^{(i)}_t$ the
advantage from the value head, $\operatorname{clip}$ the clipping operator, and
$\epsilon$ the PPO clipping range.  No
joint reward or critic is needed.  Only the shared weights tie the fleet.
Training samples one actor per episode, so the policy learns to drive different
actors, and at inference the manager drives all selected actors with it.

\begin{figure}[h]
	\centering
	\includegraphics[width=\textwidth]{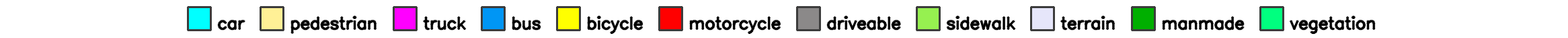}\\[2pt]
	\subcaptionbox{Scripted aggressive, aiming at the ego}{%
		\includegraphics[width=\textwidth]{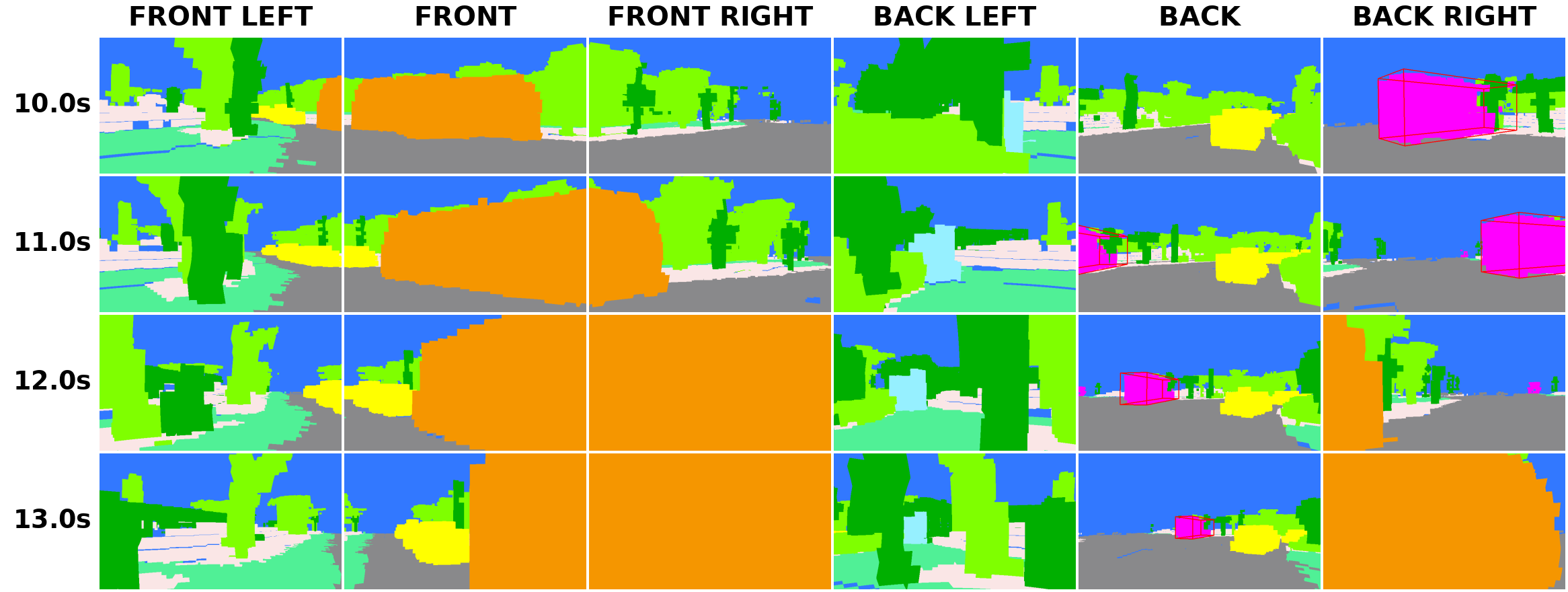}}\\[2pt]
	\subcaptionbox{Learned actor policy (Section~\ref{sec:rl})}{%
		\includegraphics[width=\textwidth]{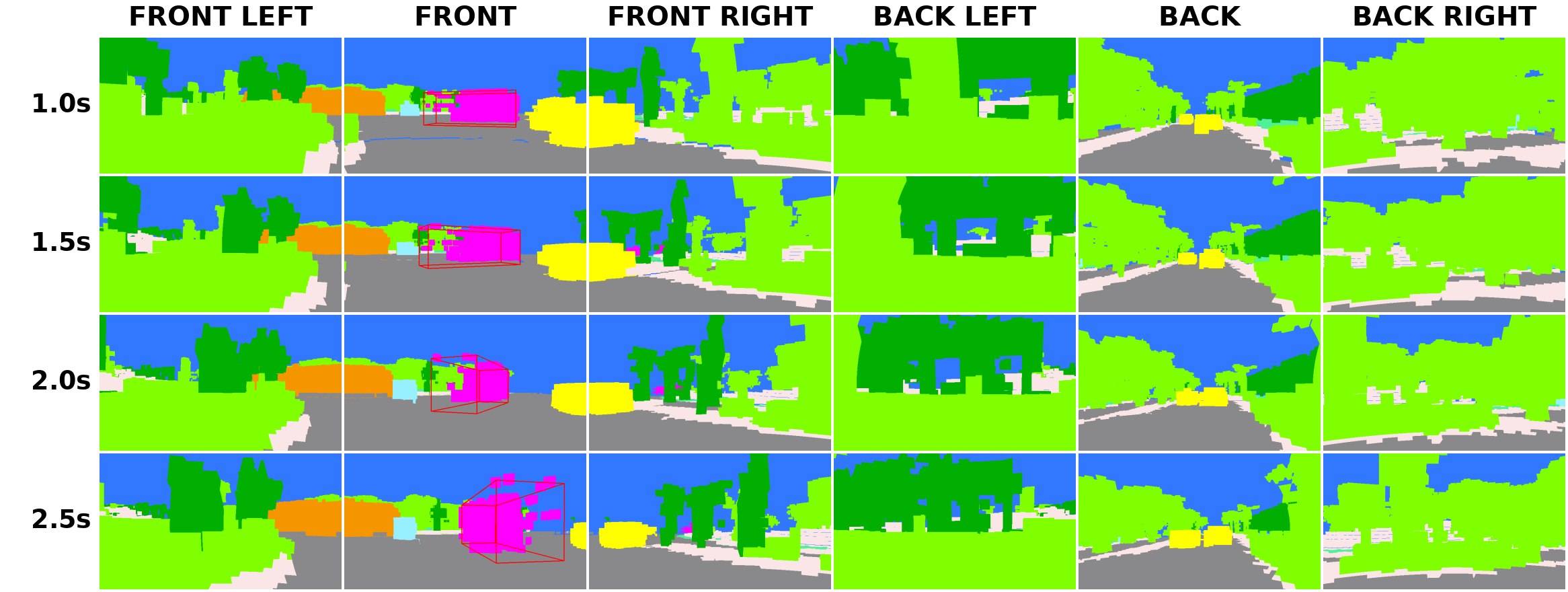}}
	\caption{The same adversarial encounter, a re-driven truck, in the ego's 2D
	semantic cameras, driven by the scripted aggressive heuristic (top) and by
	the learned actor policy (bottom).  The re-driven actor is highlighted in red
	with time along rows and the six cameras along columns.}
	\label{fig:rollout-b}
\end{figure}

The policy is trained with PPO \citep{ppo} on a single RTX\,4090, with rollouts
over eight subprocess environments.  We use Stable-Baselines3 \citep{sb3}
defaults except a $64$-step buffer, $3$ update epochs, target KL $0.15$, entropy
coefficient $0.01$, and $\gamma=0.99$.  The extractor applies single-level
deformable attention over the $50\times50$ BEV latent with $8$ heads and $4$
sampling points per head over a $128$-dim feature, and the policy and value heads
are two-layer width-$64$ MLPs over it.  The action is a normalized
$[-1,1]^2$ aim point scaled by the per-class displacement cap and tracked by the
unicycle controller.  The reward uses $r_{\text{coll}}=+20$,
$r_{\text{out}}=-10$, $\lambda=2$, $\kappa=1$.  The actor trains on $20$\,s
episodes that re-drive one uniformly sampled reachable actor per episode.  The
checkpoint is selected by the mean return over $30$ deterministic evaluation
episodes.  Training is stable, with bounded trust-region, entropy and
explained-variance curves.  Across both
schedules and all $k$, the adversarial policy's off-road and non-ego conflict
densities (Fig.~\ref{fig:actor-3metrics}) stay near the scripted aggressive
level, so its ego-threat gain is not bought by reckless driving
(Fig.~\ref{fig:rollout-b}).

\subsection{Normal Traffic Behavior}
\label{sub:app25}
The normal actor shares the architecture, observation and action space of the
adversarial policy.  Its label is the displacement toward the recorded GT pose
at the next frame, in the current ego frame and scaled by the per-class
displacement cap $d_c$, which is $1.5$--$6$\,m per frame,
\begin{equation}
	a^{\star(i)}_t=\operatorname{clip}\!\Big(\tfrac{1}{d_c(c)}\big(
	T_t^{-1}p^{\mathrm{gt}(i)}_{t+1}-p^{(i)}_t\big)\Big),
	\label{eq:bc-label}
\end{equation}
where $p^{(i)}_t$ is the driven actor position, $p^{\mathrm{gt}(i)}_{t+1}$ the
recorded next position, and $T_t$ the frame-$t\!\to\!t{+}1$ ego transform.  We train the
behavior-cloning objective of Eq.~\eqref{eq:bc-obj} on the Occ3D \emph{train}
split, for $300$ episodes, $30$ epochs, Adam lr $10^{-3}$, and batch $128$.  The
held-out action error is $\lVert a-\hat a\rVert_1{=}0.186$.  In the multi-actor
closed-loop protocol, with the GT-Following ego, both schedules, five seeds, and 150 scenes,
the clone tracks the recording to ADE $0.24$\,m against the scripted re-drive's
$0.23$\,m.  Its ego collision rate (Fig.~\ref{fig:actor-scaling}) matches the
scripted re-drive ($0.13$ vs.\ $0.11$, both run with the shared low-level
ego-obstacle brake) and stays well below the adversarial
policy, while its off-road ($0.062$ vs.\ $0.051$) and non-ego conflict ($0.334$
vs.\ $0.238$) densities stay close to the scripted re-drive
(Fig.~\ref{fig:actor-3metrics}).  It is a learned imitation of recorded traffic, not a
perfect tracker, and far less ego-aggressive than the reward-tuned adversarial
actor.

\subsection{Tokenizer Robustness to Driven Pose Deviation}
\label{sub:app18}
Adapted planners consume the frozen I$^2$-tokenizer, and since its encoder is
trained on recorded frames, a degraded token on driven frames could masquerade
as a planner failure.  We investigate this by measuring its
\emph{self}-reconstruction fidelity,
$\operatorname{decode}(\operatorname{encode}(x))$ against its own input $x$, on
the driven occupancy of the closed loop and on controlled warps of a GT frame,
over $20$ scenes (Table~\ref{tab:tokenizer-robust}).
Self-IoU/mIoU is $65.7\%$/$80.2\%$ in closed loop, $66.2\%$/$80.8\%$ under
translations, and $65.5\%$/$79.9\%$ under heading rotations. The controlled-warp scores remain close to the closed-loop value, with a
modest decline at larger rotations. This provides no evidence of a large
self-reconstruction collapse under these tested perturbations; it does not
exclude feature-distribution shift as a contributor to planner failure.

\begin{table}[h]
	\centering
	\caption{Self-reconstruction fidelity of the frozen I$^2$-tokenizer under
	controlled non-GT warping.}
	\footnotesize
	\subcaptionbox{Translation (cm)}{%
	\begin{tabular}{@{}l ccccccccc@{}}
		\toprule
		\textbf{dev (cm)} & 0 & 5 & 10 & 15 & 20 & 30 & 40 & 60 & 80\\
		\midrule
		\textbf{IoU} & 66.2 & 66.2 & 66.2 & 66.2 & 66.3 & 66.1 & 66.1 & 66.3 & 66.2\\
		\textbf{mIoU} & 80.8 & 80.8 & 80.8 & 80.8 & 81.1 & 80.9 & 80.9 & 81.2 & 81.0\\
		\bottomrule
	\end{tabular}}\\[2pt]
	\subcaptionbox{Heading (deg)}{%
	\begin{tabular}{@{}l cccccccccc@{}}
		\toprule
		\textbf{dev (deg)} & 0.0 & 0.5 & 1.0 & 1.5 & 2.0 & 3.0 & 4.0 & 5.0 & 6.0 & 8.0\\
		\midrule
		\textbf{IoU} & 66.2 & 66.1 & 66.0 & 66.0 & 66.0 & 65.9 & 65.8 & 65.7 & 65.6 & 65.5\\
		\textbf{mIoU} & 80.8 & 80.8 & 80.7 & 80.7 & 80.6 & 80.5 & 80.4 & 80.4 & 80.3 & 79.9\\
		\bottomrule
	\end{tabular}}
	\label{tab:tokenizer-robust}
\end{table}

\subsection{Adoption of Occupancy Tokenizer}
\label{sub:app19}
Among open-source tokenizers (Table~\ref{tab:tokenizer}), both the
I$^2$-tokenizer and
DOME's reconstruct strongly, at 81.2/68.3 and 83.1/77.3 mIoU/IoU.  The
selected I$^2$-tokenizer checkpoint contains $20.2$\,M weight/bias elements and
is a VQ-VAE giving deterministic observations, whereas DOME's OCC-VAE samples
its latent stochastically.  To promote reproducibility, we adopt the
I$^2$-tokenizer as the occupancy tokenizer.

\begin{table}[h]
	\centering
	\caption{Source-reported occupancy reconstruction fidelity. Parameters are
	counted from our selected checkpoint where available; unverified counts are omitted.
	Compression follows each source: OccSora includes temporal compression,
	whereas the other rows report spatial compression. OccWorld uses its default
	$50\times50$ latent configuration.}
	\footnotesize
	\resizebox{\textwidth}{!}{%
	\begin{tabular}{@{}l c c l c c c@{}}
		\toprule
		\textbf{Tokenizer} & \textbf{Year} & \textbf{Params} & \textbf{Architecture} & \textbf{Compression} & \textbf{mIoU} \rup &
		\textbf{IoU} \rup\\
		\midrule
		\rowcolor{rowlight}DOME (OCC-VAE) \citep{dome} & 2024 & --- & stochastic VAE & 64$\times$ & 83.1 & 77.3\\
		OccWorld \citep{occworld} & 2024 & --- & VQ-VAE & 16$\times$ & 66.38 & 62.29\\
		\rowcolor{rowlight}I$^2$-World \citep{i2world} & 2025 & 20.2\,M & VQ-VAE & 16$\times$ & 81.22 & 68.30\\
		OccSora \citep{occsora} & 2026 & --- & VQ-VAE & 512$\times$ & 27.4 & 37.0\\
		\bottomrule
	\end{tabular}}
	\label{tab:tokenizer}
\end{table}

\section{Benchmark and Leaderboard Details}
\label{app:details}

\subsection{Scenario, Termination, and Scoring}
\label{sub:app26}
A \emph{scenario} is one nuScenes scene under one traffic setting.  At each step
the planner observes $o_t=(B_t,\eta_t)$ of
Section~\ref{sec:overview} and emits an ego trajectory.  The scenario
ends on a collision, on a driven-ego deviation beyond $d_{\text{dev}}$ from the
reference next pose, or when the clip horizon is reached.  A collision is an
ego-footprint overlap with non-traversable occupancy above
10\% of the footprint, and sub-threshold contacts do not terminate.  DAC returns
1.0, 0.5 or 0.0 at drivable-ground footprint fractions $\ge0.5$, $\ge0.3$, or
below.  TTC extrapolates the ego at constant velocity at $+0.5$ and $+1.0$\,s
against actor boxes.  Comfort checks a rolling 6-step (3\,s) path against HUGSIM
limits \citep{hugsim}.  Eq.~\eqref{eq:pdms} uses $w_{\text{TTC}}=5$,
$w_{\text{C}}=2$.  Table~\ref{tab:planners} uses three traffic settings.  The
\emph{recorded} setting keeps actors on GT, with no re-driving and one
deterministic cell per planner, the \emph{normal} setting re-drives actors with
the \emph{learned normal actor policy} trained by behavior cloning, detailed in
Appendix~\ref{app:actor}, and the \emph{adversarial} setting re-drives them with
the \emph{learned adversarial policy} of Section~\ref{sec:rl}.  For the normal
and adversarial settings each planner runs over 150 scene starts under the
two schedules of Section~\ref{sec:eval}, \emph{spatial} ($k\in\{1,2,4\}$) and
\emph{temporal} ($k\in\{1,2,4,8,16\}$), each over seeds $s\in\{1,\dots,5\}$.  The
values in Table~\ref{tab:planners} are the mean over the five seeds, with an std
below $0.009$.

\subsection{Planner Suite}
\label{sub:app27}
\textbf{GT-Following} replays the recorded GT trajectory as its plan.
\textbf{OccWorld} \citep{occworld} jointly forecasts future occupancy and plans
autoregressively.  \textbf{Ego-MLP} \citep{bevplanner} is a perception-free
planner from the ego status alone, and \textbf{Go-Straight} a no-perception
baseline returning straight-ahead waypoints.  \textbf{BEV-Planner}
\citep{bevplanner} is a simple end-to-end planner introduced to isolate the
effect of ego status.  Its \textbf{+} and \textbf{++} variants add the ego status
to the BEV and additionally to the plan decoder over a history of frames.
\textbf{DiffusionDrive} \citep{diffusiondrive} is a truncated-diffusion
end-to-end planner that denoises a compact set of clustered trajectory anchors
conditioned on the scene.  We retain its truncated-diffusion planning head and
feed it the occupancy-token latent in place of its sparse camera features.
\textbf{ResWorld} \citep{resworld} is the temporal-residual world model
(TR-World) for end-to-end planning, with a future-guided trajectory refinement
module.  \textbf{SSR} \citep{ssr} is the navigation-guided sparse scene
representation for end-to-end planning, with a future feature predictor.
\textbf{WorldEngine (BC)} is the trajectory-vocabulary scorer of WorldEngine
\citep{li2026worldengine}, trained by behavior cloning.  \textbf{WorldEngine
(RLFT)} applies one-step REINFORCE over the trajectory vocabulary with the
imitation policy as a frozen reference \citep{li2026worldengine}.

\begingroup\revisioncolor
Table~\ref{tab:planner-bridge} summarizes which planners are used as released
versus adapted to the occupancy modality and the bridging used.

\begin{table}[h]
	\centering
	\revisioncolor
	\caption{\rev{Planner sources and modality bridging to semantic occupancy.}}
	\label{tab:planner-bridge}
	\footnotesize
	\resizebox{\textwidth}{!}{%
	\begin{tabular}{@{}l l l@{}}
		\toprule
		\textbf{Planner} & \textbf{Source} & \textbf{Modality bridging} \\
		\midrule
		OccWorld \citep{occworld} & released ckpt. & occupancy-native \\
		Ego-MLP \citep{bevplanner} & official head & ego status only (no occupancy) \\
		Go-Straight & -- & straight waypoints (no occupancy) \\
		BEV-Planner $+$/$++$ \citep{bevplanner} & adapted & frozen I$^2$-tokenizer occupancy latent replaces camera BEV features \\
		DiffusionDrive \citep{diffusiondrive} & adapted & diffusion head fed occupancy-token latent \\
		ResWorld \citep{resworld} & adapted & frozen I$^2$-tokenizer occupancy latent \\
		SSR \citep{ssr} & adapted & frozen I$^2$-tokenizer occupancy latent \\
		WorldEngine (BC/RLFT) \citep{li2026worldengine} & ported & occupancy trajectory-vocabulary scorer \\
		GT-Following & -- & human reference (no occupancy) \\
		\bottomrule
	\end{tabular}}
\end{table}
\endgroup

\subsection{Planner Reproduction}
\label{sub:app28}
OccWorld \citep{occworld} runs from its released checkpoint on the same
12-frame occupancy window, scoring its 6-frame plan under the identical
protocol.  The BEV-Planner \citep{bevplanner} family decodes per-mode
trajectories from the frozen I$^2$-tokenizer \citep{i2world} latent
and is trained with AdamW at lr $10^{-4}$ and weight decay $0.01$,
following the original setup.  Ego-MLP is the official BEV-Planner head, a
three-layer MLP over the 12-dim ego status trained by command-masked L1
regression.  Go-Straight is a simple baseline that returns six waypoints
pointing straight ahead.

\subsection{WorldEngine and RLFT Reproduction}
\label{sub:app29}
WorldEngine (BC) is our occupancy port of the WorldEngine
trajectory-vocabulary scorer \citep{li2026worldengine}.  Instead of regressing a
trajectory directly, it scores a fixed vocabulary of candidate trajectories
against the occupancy and selects the highest-scoring one, trained by behavior
cloning.  WorldEngine (RLFT) adds a
one-step REINFORCE over the vocabulary on offline train-split rollouts.  The
vocabulary holds $L{=}4096$
plans from $k$-means over the recorded train futures, and only the scoring head
is fine-tuned.  The objective combines imitation on normal frames, a
score-weighted policy gradient with importance sampling to the frozen imitation
reference, and an entropy term,
\begin{equation}
	\mathcal{L} = w_{\text{il}}\,\overline{\text{CE}(y,\log \pi)\,\mathbf{1}[\text{normal}]}
	+ w_{\text{pg}}\,\overline{-\left(\pi_\theta/\pi_{\text{orig}}\right)_{\text{detach}}\cdot s\cdot\log \pi}
	+ w_{\text{ent}}\,\overline{\textstyle\sum_k \pi\log \pi},
	\label{eq:rlft-obj}
\end{equation}
where the overbar is the mean over frames, $\pi$ is the plan-selection
distribution over the vocabulary, $y$ the imitation label, $s$ the score of the
selected plan, and $\text{CE}(\cdot,\cdot)$ the cross-entropy; the
importance-sampling weight
$(\pi_\theta/\pi_{\text{orig}})_{\text{detach}}$ is clipped at $10$ and
set to $1$ on failure frames, and $w_{\text{il}}{=}0.5$, $w_{\text{pg}}{=}1.0$,
$w_{\text{ent}}{=}0.2$, LR $5\times10^{-5}$, and $8$ epochs over $300$
train-scene rollouts.  Since
WorldEngine is evaluated on nuPlan \citep{nuplan} with the NAVSIM PDM Score
\citep{navsim}, not nuScenes/Occ3D,
its published scores are not comparable to our WorldOcc-Score.  Its method is
ported but not reproduced, so we report our port's WorldOcc-Bench scores
(Table~\ref{tab:planners}) and leave its open-loop cells empty.

\rev{We also attempted RLFT on the other benchmarked planner families; these attempts
did not succeed.  We attribute this to RLFT in a continuous action space
remaining an open question, whereas WorldEngine's discrete trajectory-vocabulary
action space appears to be the key enabler.  The RLFT row in the leaderboard is
therefore a representative demonstration rather than a claim of universality
across planner families.}

\subsection{Multi-Step RLFT}
\label{sub:app30}
The RLFT in \citep{li2026worldengine} uses a one-step return.  We also vary the TD($n$) horizon of the
policy-gradient reward, $R_t=\sum_{i=0}^{n-1}\gamma^i r_{t+i}$ with $\gamma{=}1$,
undiscounted to match the fixed-horizon driving score, computed within each
logged episode.  Here $n$ counts $0.5$\,s frames, so $n{=}6$ is $3$\,s.  All
other settings are unchanged.  Table~\ref{tab:nstep} reports the recorded-setting
score for $n\in\{1,\dots,6\}$.  The improvement peaks at $n{=}2$ ($0.309$), then
degrades for $n\ge3$ as the unnormalized return grows with $n$ while the entropy
weight is fixed.  The
degradation may reflect a scale effect: with
$\gamma{=}1$ the $n$-step return is an undiscounted sum of nonnegative per-frame
rewards whose magnitude grows with $n$ and progressively outweighs the fixed
entropy and imitation weights.  Discounting ($\gamma{<}1$) would shrink that
magnitude but truncate the effective horizon and bias the return away from the
fixed-horizon driving score, so it is not the faithful remedy.  Repeating the
sweep with a discounted return ($\gamma{=}0.99$) shows the same peak at $n{=}2$
and decline beyond, suggesting that discounting alone does not resolve the degradation.  We
hypothesize that return normalization is the next step to improve multi-step
RLFT, towards the long-horizon driving capacity.

\begin{table}[h]
	\centering
	\caption{WorldOcc-Score$_{\text{recorded}}$ versus the policy-gradient horizon
	$n$.}
	\resizebox{0.72\textwidth}{!}{%
	\begin{tabular}{@{}l cccccc@{}}
		\toprule
		\textbf{Model} & \textbf{NC} \rup & \textbf{DAC} \rup & \textbf{TTC} \rup & \textbf{Comf.} \rup & \textbf{RC} \rup & \textbf{WorldOcc-Score}$_{\text{recorded}}$ \rup\\
		\midrule
		IL base & 0.941 & 0.874 & 0.930 & 0.791 & 0.312 & 0.273\\
		\rowcolor{rowlight}RLFT $n{=}1$ & 0.976 & 0.970 & 0.969 & 0.960 & 0.331 & 0.293\\
		RLFT $n{=}2$ & \textbf{0.982} & \textbf{0.975} & \textbf{0.975} & 0.951 & \textbf{0.343} & \textbf{0.309}\\
		\rowcolor{rowlight}RLFT $n{=}3$ & 0.971 & 0.966 & 0.955 & 0.956 & 0.293 & 0.255\\
		RLFT $n{=}4$ & 0.975 & 0.965 & 0.961 & 0.955 & 0.293 & 0.259\\
		\rowcolor{rowlight}RLFT $n{=}5$ & 0.969 & 0.966 & 0.958 & 0.935 & 0.302 & 0.262\\
		RLFT $n{=}6$ & 0.976 & 0.969 & 0.960 & 0.921 & 0.293 & 0.251\\
		\bottomrule
	\end{tabular}}
	\label{tab:nstep}
\end{table}

\subsection{Per-Planner Analysis}
\label{sub:app31}
The adapted perception-based planners share the frozen I$^2$-tokenizer latent,
while OccWorld retains its native tokenizer,
and Table~\ref{tab:openloop} distinguishes source-reported metrics from our
project evaluations. The adapted ResWorld and SSR heads use the VAD
\citep{vad} collision rule, and their scores differ from the source reports.
Our adapted ResWorld variants score poorly in closed loop, OccWorld degrades the
same way despite being occupancy-native, and Ego-MLP, near-best open-loop,
degrades most under re-driven traffic.

\begin{table}[h]
	\caption{Simulator-free open-loop planning metrics, L2 in m and collision in \%.
	$^{\dagger}$OccWorld's ST-P3 protocol uses single-waypoint L2 on a 0.5\,m grid.}
	\centering
	\resizebox{\textwidth}{!}{%
	\begin{tabular}{@{}l cccc|cccc l@{}}
		\toprule
		& \multicolumn{4}{c}{L2 (m) \rdown} & \multicolumn{4}{c}{Collision (\%) \rdown} &\\
		\cmidrule(r){2-5} \cmidrule(l){6-9}
		\textbf{Method} & 1\,s & 2\,s & 3\,s & Avg & 1\,s & 2\,s & 3\,s & Avg & \textbf{Ckpt.\ source}\\
		\midrule
		OccWorld-O \citep{occworld}$^{\dagger}$ & 0.43 & 1.08 & 1.99 & 1.17 & 0.07 & 0.38 & 1.35 & 0.60 & reported\\
		OccWorld-O$^{\dagger}$ & 0.43 & 1.08 & 1.99 & 1.17 & 0.07 & 0.40 & 1.35 & 0.61 & reproduced\\
		\rowcolor{rowlight}OccWorld-O & 0.31 & 0.60 & 0.98 & 0.63 & 1.28 & 3.09 & 5.17 & 3.18 & reproduced\\
		\midrule
		BEV-Planner+ \citep{bevplanner} & 0.28 & 0.42 & 0.68 & 0.46 & 0.04 & 0.37 & 1.07 & 0.49 & reported\\
		\rowcolor{rowlight}BEV-Planner+ & 0.16 & 0.31 & 0.57 & 0.34 & 0.04 & 0.39 & 1.07 & 0.50 & reproduced\\
		BEV-Planner++ \citep{bevplanner} & 0.16 & 0.32 & 0.57 & 0.35 & 0.00 & 0.29 & 0.73 & 0.34 & reported\\
		\rowcolor{rowlight}BEV-Planner++ & 0.17 & 0.34 & 0.61 & 0.37 & 0.06 & 0.61 & 1.70 & 0.79 & reproduced\\
		Go-Straight \citep{bevplanner} & 0.38 & 0.79 & 1.33 & 0.83 & 0.15 & 0.60 & 2.50 & 1.08 & reported\\
		Go-Straight & 0.17 & 0.50 & 0.97 & 0.55 & 0.16 & 0.72 & 2.29 & 1.05 & reproduced\\
		Ego-MLP \citep{bevplanner} & 0.15 & 0.32 & 0.59 & 0.35 & 0.00 & 0.27 & 0.85 & 0.37 & reported\\
		\rowcolor{rowlight}Ego-MLP & 0.15 & 0.32 & 0.59 & 0.36 & 0.02 & 0.35 & 1.31 & 0.56 & reproduced\\
		\midrule
		ResWorld \citep{resworld} & 0.17 & 0.32 & 0.55 & 0.35 & 0.01 & 0.02 & 0.16 & 0.07 & reported\\
		\rowcolor{rowlight}ResWorld & 0.16 & 0.32 & 0.56 & 0.35 & 0.02 & 0.25 & 0.92 & 0.40 & reproduced\\
		ResWorld$\lozenge$ \citep{resworld} & 0.14 & 0.27 & 0.49 & 0.30 & 0.01 & 0.03 & 0.14 & 0.06 & reported\\
		\rowcolor{rowlight}ResWorld$\lozenge$ & 0.15 & 0.29 & 0.53 & 0.32 & 0.02 & 0.21 & 0.92 & 0.38 & reproduced\\
		SSR \citep{ssr} & 0.19 & 0.36 & 0.62 & 0.39 & 0.10 & 0.10 & 0.24 & 0.15 & reported\\
		SSR & 0.15 & 0.30 & 0.56 & 0.34 & 0.02 & 0.39 & 1.31 & 0.57 & reproduced\\
		\bottomrule
	\end{tabular}}
	\label{tab:openloop}
\end{table}

\begin{figure*}[t]
	\centering
	\includegraphics[width=\textwidth]{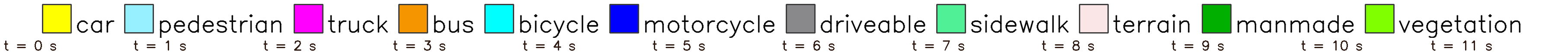}\\[2pt]
	\subcaptionbox{GT-Following, recorded traffic}{%
		\includegraphics[width=\textwidth]{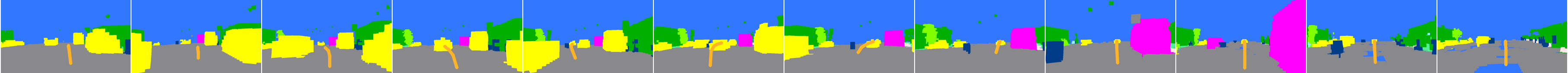}}\\[2pt]
	\subcaptionbox{GT-Following, normal traffic}{%
		\includegraphics[width=\textwidth]{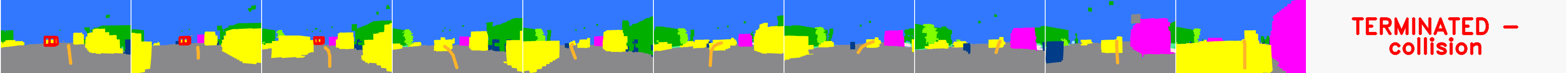}}\\[2pt]
	\subcaptionbox{GT-Following, adversarial traffic}{%
		\includegraphics[width=\textwidth]{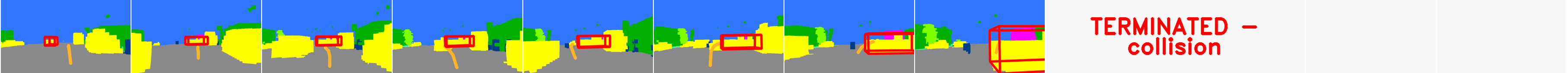}}\\[2pt]
	\subcaptionbox{Ego-MLP, recorded traffic}{%
		\includegraphics[width=\textwidth]{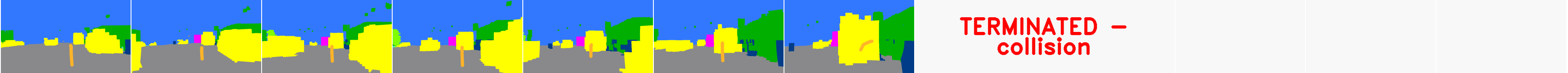}}\\[2pt]
	\subcaptionbox{Ego-MLP, normal traffic}{%
		\includegraphics[width=\textwidth]{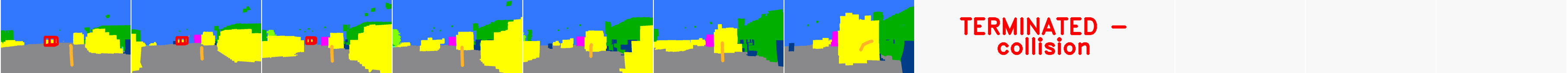}}\\[2pt]
	\subcaptionbox{Ego-MLP, adversarial traffic}{%
		\includegraphics[width=\textwidth]{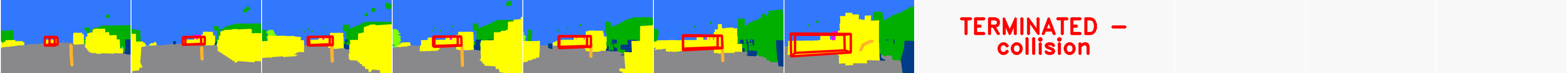}}
	\caption{Planner-by-traffic matrix: rollouts of one scene in the front-camera
	semantic view with the ego plan overlaid, showing the GT-Following oracle and
	Ego-MLP under recorded, normal, and adversarial traffic.  The re-driven actor
	is outlined in red.  Rows that terminate abnormally end with their termination
	reason, while normally-terminating rows run to the horizon cap with no banner.}
	\label{fig:rollout-a}
\end{figure*}

Table~\ref{tab:openloop} reports open-loop metrics from the papers and our runs.
The matched-protocol OccWorld and Ego-MLP L2 averages differ by at most
$0.01$\,m from their reported values. Collision rates differ, including
$0.37\%$ reported versus $0.56\%$ reproduced for Ego-MLP. ResWorld and SSR
also differ in L2 and collision; their exact reproduced values are listed rather
than assumed to match.  OccWorld's ST-P3 protocol
uses single-waypoint L2 on a 0.5\,m grid, which differs.  Under the shared per-step ADE it
scores 0.63\,m / 3.18\% instead of 1.17\,m, so the main table compares it on the
same measure as the rest.  Fig.~\ref{fig:rollout-a} shows a representative
episode in the closed loop under recorded and adversarial re-driven traffic.

\subsection{Traffic-Intensity Scaling}
\label{sub:app32}
Table~\ref{tab:scaling} reports how each planner's WorldOcc-Score degrades as
the number of re-driven actors $k$ grows, under both the normal and the
adversarial settings.  Two trends stand out.  First, the temporal schedule
degrades planners more than the spatial one at the same $k$, since the spatial
candidate pool saturates near $k\approx4$ whereas temporal encounters remain
independent and keep growing to $k=16$.  Even the GT-Following oracle falls from $0.81$ to $0.63$
under adversarial temporal traffic, so the degradation is not a planner
artifact.  Second, planners differ widely in robustness.  \rev{Ego-MLP is $0.244$ under
normal temporal traffic and $0.065$ under adversarial temporal traffic already
at $k{=}1$; it stays at $0.065$ as $k$ grows because its episodes terminate
after $8.0$ steps ($4.0$\,s) on average, with $66.6$ collisions and $76.6$
deviations per $150$ episodes (Table~\ref{tab:termination}).}  \rev{In contrast,
GT-Following reaches the time limit in $133.6/150$ episodes at $k{=}1$ and
accumulates collisions ($14.6\to50.4$) as $k$ grows, so its score falls
$0.810\to0.627$.}  \rev{Go-Straight is dominated by deviations ($114/150$) and
changes only mildly ($0.226\to0.207$).}  BEV-Planner++ and WorldEngine (RLFT)
degrade only mildly under either setting, and ResWorld stays near zero because
it already fails at $k{=}1$.  The intensity sweep thus exposes the actor load
at which each planner breaks.

\begin{table}[t]
	\centering
	\caption{WorldOcc-Score vs the number of re-driven actors $k$ under the
	normal (top) and adversarial (bottom) settings.  Spatial runs
	$k\in\{1,2,4\}$ and temporal $k\in\{1,2,4,8,16\}$.
	$^{\dagger}$GT-Following is a human reference rather than a learned planner.}
	\resizebox{0.9\textwidth}{!}{%
	\begin{tabular}{@{}l ccc ccccc@{}}
		\toprule
		& \multicolumn{3}{c}{\textbf{Spatial}} & \multicolumn{5}{c}{\textbf{Temporal}}\\
		\cmidrule(r){2-4}\cmidrule(l){5-9}
		\textbf{Planner} & $k{=}1$ & $2$ & $4$ & $k{=}1$ & $2$ & $4$ & $8$ & $16$\\
		\midrule
		\multicolumn{9}{@{}l}{\emph{Normal}}\\
		GT-Following$^{\dagger}$ & 0.842 & 0.827 & 0.815 & 0.832 & 0.831 & 0.798 & 0.770 & 0.717\\
		BEV-Planner++ \citep{bevplanner} & 0.250 & 0.251 & 0.248 & 0.252 & 0.251 & 0.248 & 0.249 & 0.246\\
		Ego-MLP \citep{bevplanner} & 0.243 & 0.242 & 0.242 & 0.244 & 0.242 & 0.242 & 0.239 & 0.235\\
		\rowcolor{rowlight}OccWorld \citep{occworld} & 0.191 & 0.190 & 0.190 & 0.194 & 0.191 & 0.190 & 0.190 & 0.189\\
		\rowcolor{rowlight}ResWorld \citep{resworld} & 0.035 & 0.035 & 0.035 & 0.035 & 0.035 & 0.035 & 0.035 & 0.035\\
		\rowcolor{rowlight}WorldEngine (RLFT) \citep{li2026worldengine} & 0.291 & 0.291 & 0.289 & 0.291 & 0.290 & 0.289 & 0.280 & 0.279\\
		\midrule
		\multicolumn{9}{@{}l}{\emph{Adversarial}}\\
		GT-Following$^{\dagger}$ & 0.814 & 0.792 & 0.767 & 0.810 & 0.791 & 0.743 & 0.679 & 0.627\\
		BEV-Planner++ \citep{bevplanner} & 0.250 & 0.245 & 0.243 & 0.249 & 0.244 & 0.243 & 0.236 & 0.225\\
		Ego-MLP \citep{bevplanner} & 0.065 & 0.065 & 0.065 & 0.065 & 0.065 & 0.065 & 0.064 & 0.064\\
		\rowcolor{rowlight}OccWorld \citep{occworld} & 0.188 & 0.188 & 0.187 & 0.191 & 0.188 & 0.182 & 0.177 & 0.168\\
		\rowcolor{rowlight}ResWorld \citep{resworld} & 0.035 & 0.035 & 0.035 & 0.035 & 0.035 & 0.035 & 0.034 & 0.034\\
		\rowcolor{rowlight}WorldEngine (RLFT) \citep{li2026worldengine} & 0.284 & 0.279 & 0.276 & 0.284 & 0.271 & 0.256 & 0.239 & 0.212\\
		\bottomrule
	\end{tabular}}
	\label{tab:scaling}
\end{table}

\begin{table}[t]
	\centering
	\revisioncolor
	\caption{Per-150-episode termination statistics under adversarial temporal
	traffic (mean over 5 seeds).  Counts sum to 150; mean steps at 0.5\,s/step;
	score is normalized WorldOcc-Score.}
	\label{tab:termination}
	\begin{tabular}{@{}l r r r r r r@{}}
		\toprule
		\textbf{Planner} & $k$ & \textbf{Coll.} & \textbf{Dev.} & \textbf{T.L.} &
		\textbf{Mean steps} & \textbf{Score}\\
		\midrule
		GT-Following & 1 & 14.6 & 1.8 & 133.6 & 37.2 & 0.810\\
		GT-Following & 2 & 18.8 & 2.0 & 129.2 & 36.8 & 0.791\\
		GT-Following & 4 & 32.4 & 1.8 & 115.8 & 35.5 & 0.743\\
		GT-Following & 8 & 42.8 & 2.0 & 105.2 & 33.5 & 0.679\\
		GT-Following & 16 & 50.4 & 1.8 & 97.8 & 31.8 & 0.627\\
		\midrule
		Ego-MLP & 1 & 66.6 & 76.6 & 6.8 & 8.0 & 0.065\\
		Ego-MLP & 2 & 66.8 & 76.2 & 7.0 & 7.9 & 0.065\\
		Ego-MLP & 4 & 66.8 & 76.4 & 6.8 & 7.9 & 0.065\\
		Ego-MLP & 8 & 66.2 & 77.0 & 6.8 & 8.0 & 0.064\\
		Ego-MLP & 16 & 65.4 & 78.4 & 6.2 & 7.9 & 0.064\\
		\midrule
		Go-Straight & 1 & 24.0 & 114.0 & 12.0 & 14.0 & 0.226\\
		Go-Straight & 2 & 23.4 & 114.4 & 12.2 & 14.1 & 0.226\\
		Go-Straight & 4 & 25.0 & 112.4 & 12.6 & 14.1 & 0.221\\
		Go-Straight & 8 & 25.8 & 112.0 & 12.2 & 13.8 & 0.216\\
		Go-Straight & 16 & 25.6 & 112.2 & 12.2 & 13.8 & 0.207\\
		\bottomrule
	\end{tabular}
\end{table}

\begin{figure}[h]
	\centering
	\resizebox{\textwidth}{!}{\input{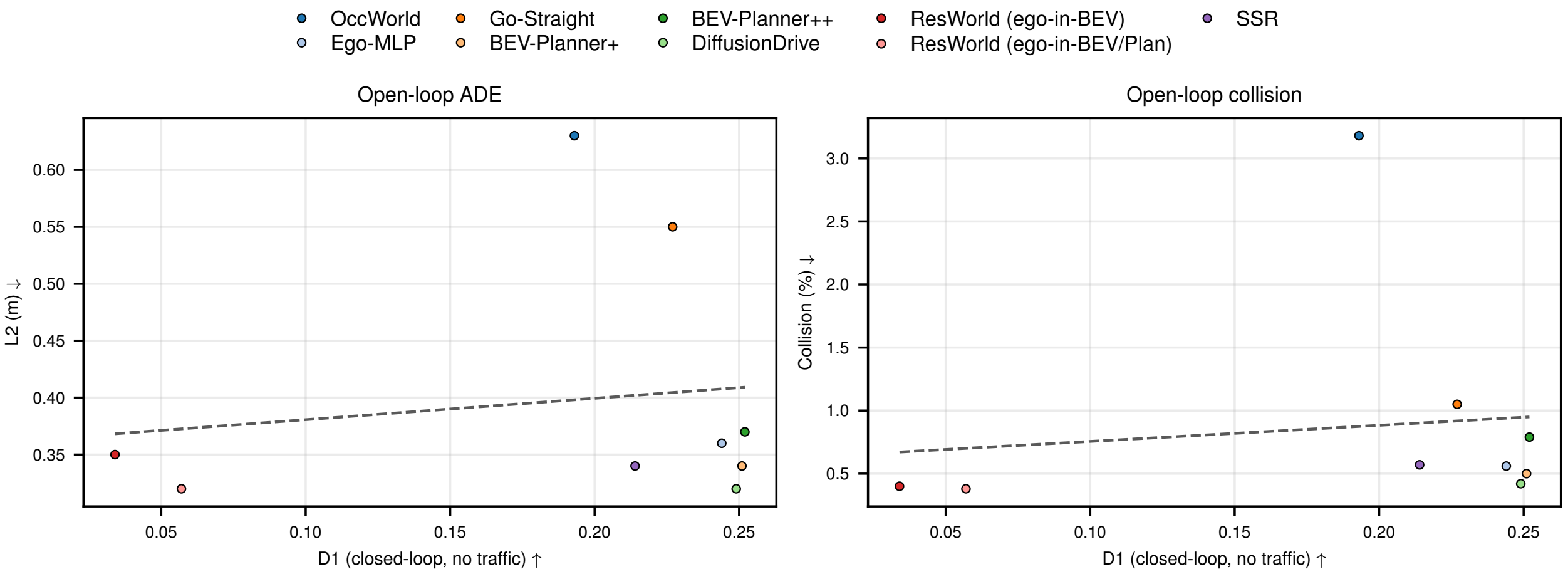}}
	\caption{Open-loop and closed-loop correlation across nine planner variants. GT-Following and WorldEngine variants are excluded because they are a reference or lack the open-loop measures.}
	\label{fig:d1-correlation}
\end{figure}

\section{\rev{Scope and Discussion}}
\label{app:scope}
\begingroup\revisioncolor

This appendix collects the scope questions that the main text raises, with
the full protocols and per-sample results.

\paragraph{Occupancy-Level vs.\ Sensor-Level Evaluation.}
WorldOcc-Bench evaluates planners on semantic occupancy.  This is the native
interface of occupancy-based planners such as OccWorld \citep{occworld} and
OccLLaMA \citep{occllama}, which plan directly on an occupancy state.

The simulator supplies GT occupancy, the recorded state used as the reference.
This provides a controlled reference for planning with annotated occupancy.
Predicted occupancy introduces perception error, and compatible planners can
consume it through the same interface. GT-input scores are not a mathematical
upper bound for a fixed learned planner: input errors can change its behavior
in either direction. A matched GT-versus-predicted comparison would be needed
to measure that perception gap; this benchmark does not perform that comparison.

The benchmark does not model perception errors or sensing uncertainty itself.
Its scores are a controlled, planner-centric complement to sensor-level
benchmarks, not a substitute for them.  Extending it toward sensor integration
is tractable because occupancy is a unified intermediate representation.
Occupancy-centric generators such as UniScene \citep{uniscene}, UniScene-V2
\citep{uniscenev2}, and AnyScene \citep{anyscene} synthesize occupancy and
bridge it to camera and LiDAR.  Rendering observations from the occupancy
state, or composing WorldOcc with such a generator, would extend coverage
beyond recorded scenes. The transition reconstructs each state from its
same-time reference; its conditional label-difference bound
(Appendix~\ref{app:overlap}) does not guarantee low counterfactual error or
bounded cumulative pose drift.

\begin{table}[t]
	\centering
	\revisioncolor
	\caption{\rev{Capability comparison between OccSim \citep{occsim} and WorldOcc.
	\cmark{} indicates supported or claimed; \xmark{} indicates not supported or
	not claimed.}}
	\label{tab:occsim-compare}
	\footnotesize
	\begin{tabular}{@{}l cc@{}}
		\toprule
		\textbf{Capability} & \textbf{OccSim \citep{occsim}} & \textbf{WorldOcc (ours)} \\
		\midrule
		Real-time operation (FPS) & \xmark\ (1.47\,FPS) & \cmark\ (ego operator: $\sim$1270\,FPS) \\
		Long-horizon rollouts & \cmark\ (3000+ frames) & \cmark\ (16\,s) \\
		Novel-scene / multi-kilometer synthesis & \cmark & \xmark \\
		HD-map- and log-free generation & \cmark & \xmark \\
		Learnable normal/adversarial actor policy & \xmark & \cmark \\
		Standardized closed-loop planner benchmark & \xmark & \cmark \\
		Planner post-training (RLFT) & \xmark & \cmark \\
		\bottomrule
	\end{tabular}
\end{table}

\paragraph{Diverse Traffic and Interaction Patterns.}
``Diverse traffic'' here refers to the behavior of traffic participants.
Controlled actors are re-driven by a learnable policy, and road topology can be
varied by switching the dataset.  Weather is outside the occupancy modality.
Table~\ref{tab:interaction-stats} reports the interaction-pattern statistics,
and Table~\ref{tab:conflict-kinematics} reports the corresponding kinematic
envelope.  \rev{For the approach angle, minimum TTC, and approach speed, $n$
counts tracks with a finite value; the three metrics require non-zero actor
motion, positive closing speed, and positive approach, respectively.}

\begin{table}[h]
	\centering
	\revisioncolor
	\caption{\rev{Interaction-pattern statistics over 150 closed-loop scenes at
	$2\,$Hz (median [interquartile range]).  $n$ counts tracks with a finite
	value.}}
	\label{tab:interaction-stats}
	\footnotesize
	\begin{tabular}{@{}l c c@{}}
		\toprule
		\textbf{Metric} & \textbf{Learned adversarial} & \textbf{Scripted aggressive} \\
		\midrule
		Approach angle (deg) & 35.0 [15.0, 85.4] ($n{=}104$) & 29.5 [2.8, 65.9] ($n{=}81$) \\
		Min.\ TTC (s) & 1.69 [0.76, 5.60] ($n{=}94$) & 1.75 [1.02, 4.22] ($n{=}107$) \\
		Lateral offset (m) & 2.87 [1.73, 6.30] ($n{=}130$) & 3.37 [1.16, 7.25] ($n{=}130$) \\
		Approach speed (m/s) & $5.10$ [2.77, 8.09] ($n{=}99$) & $5.95$ [3.15, 8.75] ($n{=}110$) \\
		\bottomrule
	\end{tabular}
\end{table}

\rev{These descriptive kinematic statistics are not a verification of comfort or
physical feasibility for every trajectory; the lateral-acceleration upper
quartile is $4.97\,$m/s$^2$.}

\begin{table}[h]
	\centering
	\revisioncolor
	\caption{\rev{Kinematic envelope over the same 150 scenes: median [IQR] of
	per-track 95th percentiles over 130 actor tracks.}}
	\label{tab:conflict-kinematics}
	\footnotesize
	\begin{tabular}{@{}l c@{}}
		\toprule
		\textbf{Quantity} & \textbf{Value} \\
		\midrule
		Actor speed (m/s) & 5.03 [2.73, 6.82] \\
		$|a_{\text{lon}}|$ (m/s$^2$) & 2.47 [1.17, 3.83] \\
		$|a_{\text{lat}}|$ (m/s$^2$) & 2.32 [1.11, 4.97] \\
		$|\dot{\psi}|$ (deg/s) & 42.3 [24.6, 74.1] \\
		\bottomrule
	\end{tabular}
\end{table}

\paragraph{Relation to OccSim.}
OccSim \citep{occsim} is a long-horizon 3D occupancy simulator driven by a
world model.  Conditioned on a single initial frame and a sequence of future
ego actions, it generates over 3000 continuous frames and multi-kilometer maps
without HD maps or recorded logs, a stable rollout over $80\times$ longer than
prior occupancy world models.  It combines a W-DiT static occupancy world model
with a Layout Generator that learns agent layouts and populates the dynamic
foreground with reactive agents.  Its downstream use is to synthesize
large-scale occupancy data for pre-training 4D forecasting models; its reported
zero-shot performance is measured relative to an in-domain upper bound.  Its reported limitations include heuristic map
fusion and lane-graph extraction.

The two systems answer different questions.  Real-time driving simulation
generally needs about 10\,FPS to match a 10\,Hz sensor rate; HUGSIM
\citep{hugsim}, a real-time simulator, reports about 89\,FPS.  WorldOcc's GPU-resident Ego Warp-Recenter operator runs at
about 1270\,FPS, excluding controller, actor editing, tokenizer and planner
inference, while OccSim reports 1.47\,frames/s in its Table~3. These timings
cover different pipelines and are not an end-to-end speedup comparison.  WorldOcc
is reference-anchored and uses no learned world model, and it targets
closed-loop occupancy-planner evaluation with learnable normal and adversarial
traffic (WorldOcc-Bench).  OccSim generates at scale; WorldOcc evaluates
planners with an efficient occupancy transition.

\paragraph{Relation to Replay and Reactive Simulators.}
We do not compare with replay and reactive simulators because they do not
expose Occ3D-style semantic occupancy (Table~\ref{tab:baseline-modality}).
NAVSIM \citep{navsim}, nuPlan \citep{nuplan}, and Waymax \citep{waymax} operate
on camera observations, object boxes, or HD maps.  GPUDrive \citep{gpudrive} is
an object-centric multi-agent simulator on Waymo Open Motion.  MetaDrive
\citep{metadrive} and ScenarioNet \citep{scenarionet} are synthetic or replay
simulators with scenario-level representations.  Their protocols target driving
score, infractions, or object-level forecasting, which are different quantities
from occupancy fidelity.  An occupancy-fidelity comparison is therefore
undefined, and comparing image- or object-level fidelity with voxel-level IoU
would not be meaningful.  We compare on the shared occupancy axis against
occupancy world models (Table~\ref{tab:fidelity}) and cite these systems as
related work.

\begin{table}[H]
	\centering
	\revisioncolor
	\caption{\rev{Observation modality of the replay and reactive simulators.
	None exposes Occ3D-style semantic occupancy, so an occupancy-fidelity
	comparison is undefined.}}
	\label{tab:baseline-modality}
	\footnotesize
	\resizebox{\textwidth}{!}{%
	\begin{tabular}{@{}l l c@{}}
		\toprule
		\textbf{Method} & \textbf{Observation / representation} & \textbf{Semantic occupancy} \\
		\midrule
		NAVSIM \citep{navsim} & Camera observations & \xmark \\
		nuPlan \citep{nuplan} & Camera/LiDAR/radar + HD map & \xmark \\
		Waymax \citep{waymax} & Object boxes + roadgraph (no sensors) & \xmark \\
		GPUDrive \citep{gpudrive} & Object-centric multi-agent state (Waymo Open Motion) & \xmark \\
		MetaDrive \citep{metadrive} & Synthetic scenarios, configurable sensors & \xmark \\
		ScenarioNet \citep{scenarionet} & Real-world scenarios + HD map + object annotations & \xmark \\
		\midrule
		WorldOcc (ours) & Semantic occupancy (Occ3D) & \cmark \\
		\bottomrule
	\end{tabular}}
\end{table}
\begin{table}[H]
	\centering
	\revisioncolor
	\caption{\rev{Generator-composition fidelity on CARLA-SC (16 frames,
	1.6\,s); IoU/mIoU against the reference stream, mean over ten samples.}}
	\label{tab:gen-compose}
	\footnotesize
	\begin{tabular}{@{}l c c@{}}
		\toprule
		\textbf{Reference stream} & \textbf{IoU} & \textbf{mIoU} \\
		\midrule
		DynamicCity sample 1 & 0.984 & 0.932 \\
		DynamicCity sample 2 & 0.975 & 0.958 \\
		DynamicCity sample 3 & 0.984 & 0.925 \\
		DynamicCity sample 4 & 0.986 & 0.948 \\
		DynamicCity sample 5 & 0.981 & 0.950 \\
		DynamicCity sample 6 & 0.994 & 0.966 \\
		DynamicCity sample 7 & 0.987 & 0.955 \\
		DynamicCity sample 8 & 0.984 & 0.903 \\
		DynamicCity sample 9 & 0.979 & 0.945 \\
		DynamicCity sample 10 & 0.981 & 0.952 \\
		DynamicCity mean $\pm$ std & 0.983 $\pm$ 0.005 & 0.943 $\pm$ 0.018 \\
		Recorded (CARLA-SC) & 0.988 & 0.958 \\
		\midrule
		Gap (recorded $-$ mean) & 0.004 & 0.014 \\
		\bottomrule
	\end{tabular}
\end{table}

\begingroup\setlength{\intextsep}{6pt}
\begin{table}[H]
	\centering
	\revisioncolor
	\caption{\rev{Transition consistency on the val split of NuPlan-Occ mini
	(30 scenes); Lane IoU measures the projected nuPlan lane boundaries.}}
	\label{tab:nuplan-occ}
	\footnotesize
	\begin{tabular}{@{}l ccc@{}}
		\toprule
		\textbf{Horizon} & \textbf{IoU} & \textbf{mIoU} & \textbf{Lane IoU} \\
		\midrule
		$1\,$s  & 0.9015 & 0.6697 & 0.8689 \\
		$5\,$s  & 0.8905 & 0.6359 & 0.8522 \\
		$10\,$s & 0.8814 & 0.6166 & 0.8394 \\
		$20\,$s & 0.8805 & 0.6248 & 0.8307 \\
		\bottomrule
	\end{tabular}
\end{table}
\endgroup

\paragraph{Lane-Level and Traffic-Rule Signals.}
Semantic occupancy already encodes drivable, sidewalk, and road classes; the map
adds lane boundaries, traffic-light states, and right-of-way.  A rule module
maps these to actor behavior and planner constraints without altering the
transition, and HDMap-based closed-loop simulation is well studied
\citep{nuplan,carla}.  \rev{The main benchmark is defined on Occ3D-nuScenes, which
carries no traffic-light states.  The NuPlan-Occ check in Section~5 scores compliance
on NuPlan-Occ mini; extending the benchmark to signal-aware planners is future
work.  All NuPlan-Occ numbers in Section~5 and
Table~\ref{tab:nuplan-occ} use 30 recorded scenes of the val split of NuPlan-Occ mini, drawn with seed 0
from the 218 scenes with at least 100 frames (sorted by token) and truncated to
$20\,$s each.}

\endgroup

\end{document}